\documentclass[11pt]{article}
\usepackage[utf8]{inputenc}
\usepackage[T1]{fontenc}
\usepackage{graphicx}
\usepackage{caption}
\usepackage{authblk}
\usepackage{subcaption}
\usepackage[section]{placeins}
\usepackage{hyperref}
\usepackage{MnSymbol}
\usepackage[backend=biber,style=ieee]{biblatex}
\usepackage{lscape} % this package is for landscape table 1 
\usepackage{adjustbox}
\usepackage[nottoc]{tocbibind}
\usepackage{setspace}
\usepackage{rotating}
\graphicspath{{Fig/}}

\DeclareUnicodeCharacter{00C7}{\c C}
\DeclareUnicodeCharacter{0301}{\'{e}}

\begin{document}

\title{Literature Review of the Recent Trends and Applications in various Fuzzy Rule based systems}

\author{Ayush K. Varshney$^{*}$, Vicen\c{c} Torra \\
Department of Computing sciences\\
Umeå University, Sweden \\
Emails: varshneyayush90@gmail.com$^{*}$, vtorra@ieee.org
}
\date{}
\maketitle
\vspace*{-0.8cm}
\textbf{Abstract: }Fuzzy rule based systems (FRBSs) is a rule-based system which uses linguistic fuzzy variables as antecedents and consequent to represent human understandable knowledge. They have been applied to various applications and areas throughout the soft computing literature. However, FRBSs suffers from many drawbacks such as uncertainty representation, high number of rules, interpretability loss, high computational time for learning etc. To overcome these issues with FRBSs, there exists many extensions of FRBSs. This paper presents an overview and literature review of recent trends on various types and prominent areas of fuzzy systems (FRBSs) namely genetic fuzzy system (GFS), hierarchical fuzzy system (HFS), neuro fuzzy system (NFS), evolving fuzzy system (eFS), FRBSs for big data, FRBSs for imbalanced data, interpretability in FRBSs and FRBSs which use cluster centroids as fuzzy rules. The review is for years 2010-2021. This paper also highlights important contributions, publication statistics and current trends in the field. The paper also addresses several open research areas which need further attention from the FRBSs research community.

\maketitle

\section{Introduction}
Rule-based Systems \cite{durkin4expert} are generally used to model human problem-solving activity and behavior using classical IF-THEN rules. When the antecedent(s) of the rule is satisfied, then the rule is fired. Classical approaches deal with the bivalent logic, but it fails to cover the imprecision and uncertainty present in the knowledge representation. Fuzzy sets \cite{zadeh1965information} introduced by Zadeh are widely used to handle uncertainties and imprecision. Fuzzy sets with rule-based systems leads to Fuzzy Rule Based Systems (FRBSs, see Table \ref{abbreviation} for further abbreviation and their full name) \cite{zadeh1973outline, mamdani1974application}. FRBSs considers fuzzy statements as antecedents and consequents, they can handle uncertainty and are more robust compared to the classical rule-based system. They use Linguistic Variables \cite{zadeh1975concept} to represent features in antecedents, whose values are context dependent on the membership function of the feature. Rules generated by FRBSs are human interpretable\footnote{In this article, we have considered interpretable and explainable as equivalent since they both imply human understandability} and can be used to understand how the system works. A typical rule in FRBSs looks like:

\vspace{0.2cm}$Rule \; R_i: If A_1 \; is \; x_{i1} \; \text{and} \; A_2 \; is \; x_{i2} \; \text{and} \; ...\; \text{and}\; A_n\; is\; x_{in}\; \text{then output is } O_j$

\vspace{0.2cm}here, $R_i$ is the $i^{th}$ rule in the rulebase; $A_1,\; A_2,\; ..., A_n$ are the features in a dataset; $x_{i1},\; x_{i2},\;...\;x_{in}$ are the linguisitic variables and $O_j$ is the output variable. 

\begin{figure}
     \centering
     \begin{subfigure}[b]{0.45\textwidth}
         \centering
         \includegraphics[width=\textwidth]{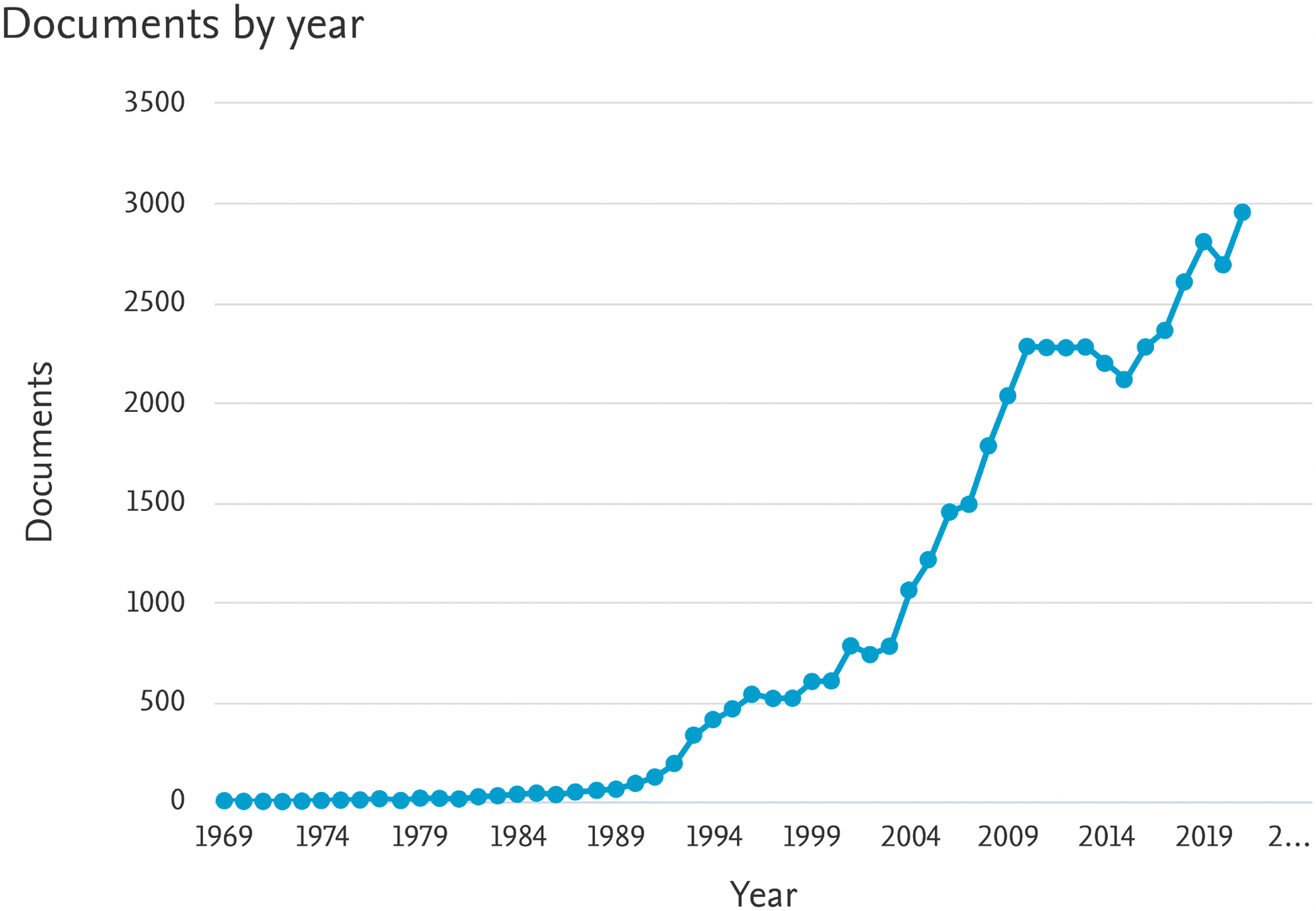}
         \caption{ }
         \label{fuzz publications 2010-2021}
     \end{subfigure}
     \hfill
     \begin{subfigure}[b]{0.45\textwidth}
         \centering
         \includegraphics[width=\textwidth]{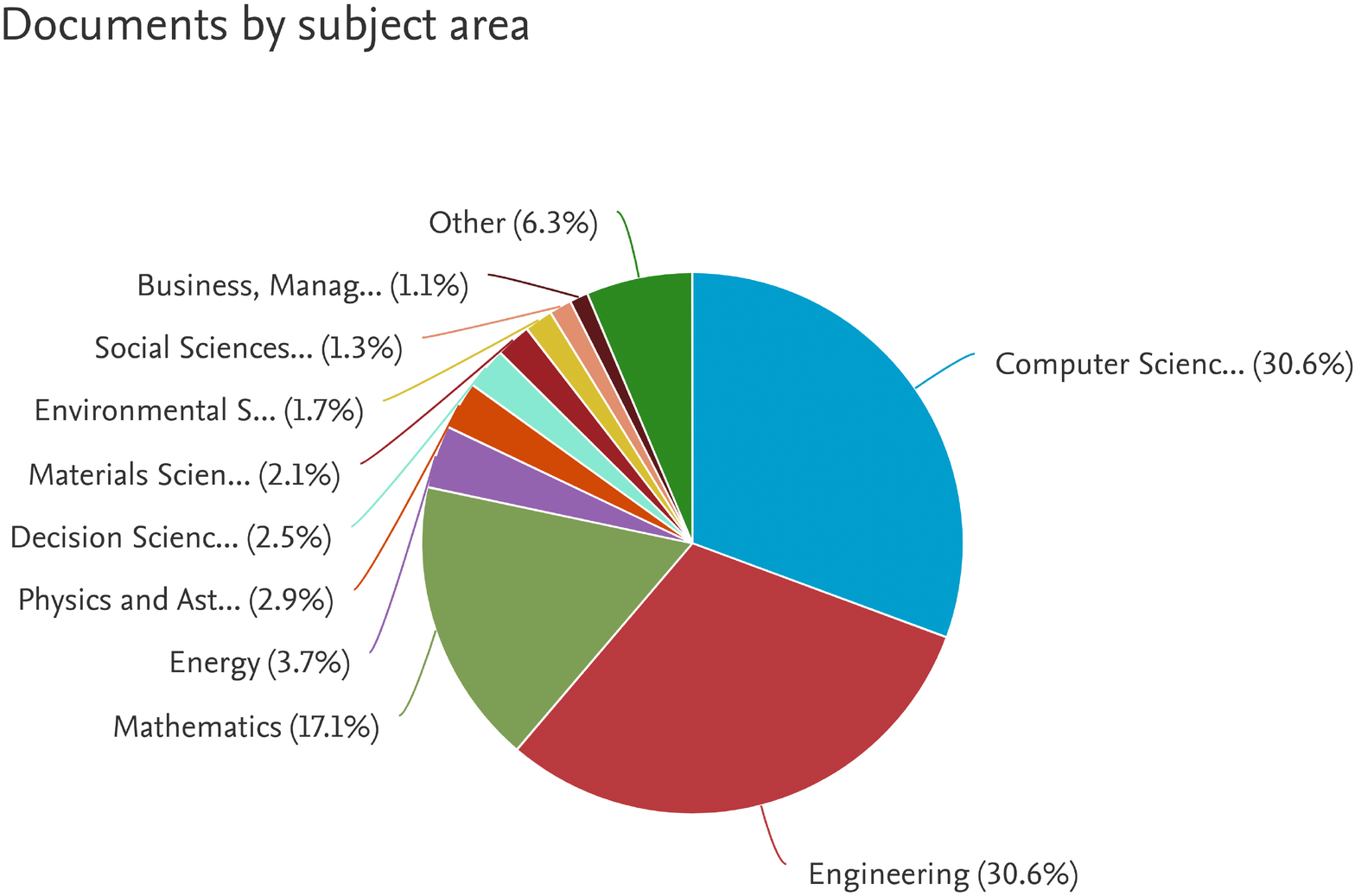}
         \caption{ }
         \label{Area-wise publications in FRBSs}
     \end{subfigure}
     
    \caption{(a) Number of published articles in the area of fuzzy systems throughout the years (up to 2021) (b) Area-wise categorization for the published articles.}
    \label{FRBSs_introduction}
\end{figure}

FRBSs (or fuzzy system(s)) have been used for more than 40 years in many areas of computer science and engineering, there exists 45,287 articles on fuzzy system(s) in the Scopus database (\url{https://www.scopus.com/}). Scopus has a comprehensive data coverage. In case of fuzzy systems, Scopus has the repository of papers from all the major publication venues such as IEEE Transactions on Fuzzy Sets and Systems (TFS), WCCI/IEEE Fuzzy conference, Fuzzy Sets and Systems (FS\&S) and many more. Despite little fluctuation, the number of articles published in the area of fuzzy systems is constantly growing which shows the growing attention of research community in FRBSs (see Fig. \ref{fuzz publications 2010-2021}). Fig. \ref{Area-wise publications in FRBSs} shows the area-wise percentage of the published articles in fuzzy systems highlighting the major area of research in FRBSs. 

\begin{table}[!ht]
    \centering
    \begin{adjustbox}{width=\textwidth}
        \begin{tabular}{l|l|l|l}
        \hline
        \hline
            Abbreviation & Full name & Abbreviation & Full name \\
            \hline
            DNN & Deep neural network & eTS & Evolving Takagi Sugeno \\
            \hline
            eFS & Evolving Fuzzy system & EAa & Evolutionary Algorithms \\
            \hline
            FRBSs & Fuzzy Rule Based Systems & FIS & Fuzzy Inference System \\
            \hline
            FM & Fuzzy Modelling & GFS & Genetic Fuzzy System \\
            \hline
            GAs & Genetic AlgorithmsHFS & Hierarchical Fuzzy System & HFS \\
            \hline
            LVs & Linguistic variables & KB & Knowledge base \\
            \hline
            MOEA & Multi-objective Evolutionary Algorithm & MFs & Membership functions \\
            \hline
            NFS & Neuro Fuzzy System & NN & Neural Network \\
            \hline
            PSO & Particle Swarm Optimization & RB & Rule base \\
            \hline
            RQ & Research Question & TSK & Takagi-Sugeno-Kang\\
        \hline
        \hline
        \end{tabular}
    \end{adjustbox}
    \caption{Used abbreviation and their full name.}
    \label{abbreviation}
\end{table}

The main objective of this article is to provide the literature review of recent trends and applications in the FRBSs during the years 2010-2021. There exists many literature review in the literature such as the following: Jang et. al \cite{jang1993anfis} reviews the structure and parameter optimization for Neuro Fuzzy System (NFS) which was further considered in \cite{kosko1998neural} and \cite{karaboga2019adaptive}; terminologies and usage of Genetic Fuzzy System (GFS) has been reviewed in \cite{fernandez2019evolutionary}; Hierarchical Fuzzy System (HFS) and their advances were reviewed in \cite{torra2002review} and \cite{di2006survey}; online learning in fuzzy systems was reviewed in \cite{angelov2008evolving} and \cite{leite2020overview}; a literature review of three decades for FRBSs, GFS, NFS and HFS has been given in \cite{ojha2019heuristic}; Explainability perspective of FRBSs has been reviewed in \cite{moral2021explainable}. In this article, all the major extensions and areas where FRBSs have been used in the past 12 years such as FRBSs in big data, imbalanced data, cluster centroids as rule have been considered. These topics are not considered in any other review articles. This work can be considered as an extension of \cite{ojha2019heuristic} and \cite{moral2021explainable} in the sense that it identify the current trends and applications related to all the major areas of FRBSs based on the number of published articles and the citations related to them during 2010-2021. This paper has considered original research articles related to fuzzy systems and its various types written in English, published in the period 2010 to 2021. 

In this paper, we have identified 8 key areas where FRBSs have contributed significantly and present a systematic literature review. These areas are Genetic/ Evolutionary Fuzzy Systems (GFS/EFS) \cite{herrera1996adaptation}, Hierarchical Fuzzy Systems (HFS) \cite{gegov1995decomposition}, Adaptive Neuro Fuzzy Inference System \cite{jang1991fuzzy} (ANFIS/NFS), Evolving Fuzzy Systems (eFS) \cite{kasabov2001evolving}, FRBSs for Big Data \cite{robles2009evolutionary}, FRBSs for Imbalanced Datasets \cite{batuwita2010fsvm}, Clusters centroids as rules in FRBSs \cite{yager1994generation} and FRBSs for Interpretability/Explainability \cite{hagras2018toward}. 

A GFS uses genetic/evolutionary approaches to improve the learning ability of FRBSs. A HFS solves the curse of dimensionality for FRBSs, NFS improves the approximation ability of FRBSs using neural networks and dynamic systems for streaming data is considered in eFS. FRBSs are seen as good solutions for big data and imbalanced data. In the recent years the interpretability in FRBSs has gained popularity due to high dimensional and big data and rules are initialized with cluster centroids to limit the number of rules in FRBSs. This article aims to find the answers to the following research question (RQ):

\begin{itemize}
    \item \textit{RQ1:} Which are the fundamental concepts in FRBSs and its variants? \\  Motivation: This is needed to understand the basic knowledge regarding FRBSs and its types.
    \item \textit{RQ2:} What are the publishing statistics associated with each type of FRBSs? \\Motivation: These statistics can help the researchers to identify the direction of growth in the field. These statistics can also help in identifying the publication venues.
    \item \textit{RQ3:} What are the trending areas for each type of FRBSs? \\ Motivation: Identify the trending areas can help to further enhance the FRBSs, as it is a key issue for primary as well as for secondary researchers.
    \item \textit{RQ4:} What are the open problems in FRBSs? \\ Motivation: Identifying the key problems in FRBSs can help researchers in inducing the state of the art of the field. 
    
\end{itemize}

The paper is organized as follows: Section II describes the classical FRBSs with its variants; Section III provides the literature review for recent trends in the above-mentioned areas along with the details on the selection procedures, Section IV provides current research trends along with open problems for fuzzy systems, and Section V presents the conclusion. 

\section{Fuzzy Rule Based Systems (FRBSs):} \label{FRBS}

\begin{figure}
    \centering
    \includegraphics[width=\textwidth]{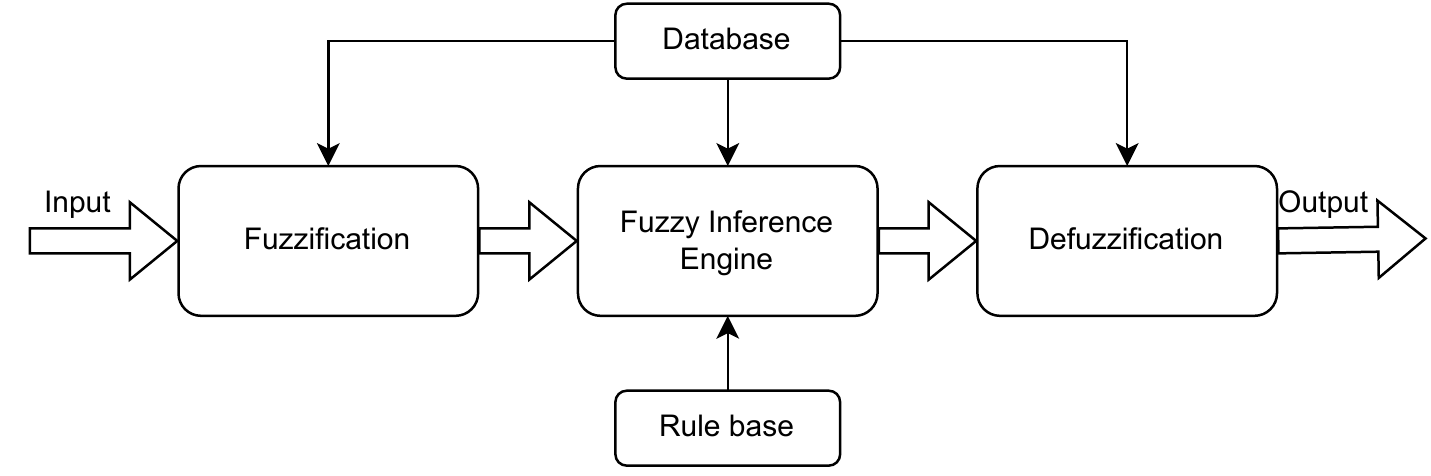}
    \caption{General structure of Fuzzy Rule Based Systems.}
    \label{FRBSs_generic}
\end{figure}

Classical fuzzy systems are based on Mamdani \cite{mamdani1974application} approach. Its generic structure is shown in Fig. \ref{FRBSs_generic}. The fuzzification module establishes a mapping between real-valued input data to fuzzy values based on some membership function. Similarly, the defuzzification module establishes a mapping between fuzzy values to real-valued output domain. The data base contains the linguistic variables (LV) and the specific membership function associated with them. Rules are the usual way to organize knowledge in natural language. Usually, Database and Rule base are part of the Knowledge base (KB). Mamdani’s rules involves the use of LVs in antecedent and consequent. Rules are represented by the set of linguistic variables and an output associated with them. I.e., a rule can have multiple inputs and a single output. The rule base contains the set of rules for the specific application. Rule base can be represented in many ways. Rule base either has a list of rules or a decision table which is a compact representation of rules. The fuzzy inference engine infers the fuzzy output from the inputs according to the rules generated in the Rule base.

Mamdani's FRBSs suffer from some drawbacks which were first highlighted in \cite{bastian1994handle, carse1996evolving} due to the use of LVs \cite{moral2021explainable}:
\begin{enumerate}
    \item Lack of flexibility due to rigid partition of input and output space.
    \item It is difficult to find a fuzzy partition when input variables are mutually dependent.
    \item The homogeneous partition between input and output is inefficient and does not scale well.
    \item The size of a KB increases rapidly with the increase in the number of variables and linguistic terms for each variable.
\end{enumerate}
	
There exist multiple variants of the classical fuzzy rule-based system which addresses these issues in FRBSs:
\begin{enumerate}
    \item Mamdani FRBSs with input output scaling: 
    Transforming the inputs and outputs to introduce more flexibility was introduced in \cite{procyk1979linguistic}. The authors introduce a simple linear scaling function of the form below:
    \[f(z)=\lambda \cdot z+c\]
    Here $\lambda$ is the scaling factor and $c$ is the offset for the variable $z$. Non-linear scaling is also possible, which may be of the form:
    
    \[ f(z)=sign(z)+|z|^\alpha \]
    
    Non-linear scaling factor ($\alpha$) is used to control the sensitivity in the origin region. This way each n-dimensional rule in the FRBSs will be of the form:
    
    $\text{IF} \; f(x_1) \; is \; l_{1i} \; \text{and} \; f(x_2) \; is \; l_{2i} \; \text{and} \; ... \; \text{and} \; f(x_n) \; is \; l_{ni} \; then \; y \; is \; Y$
    
    where $l_{ij}$ is the $j^{th}$ linguistic variable for the $i^{th}$ feature.
    
    \item DNF Mamdani Fuzzy rule-based system \cite{gonzalez1994learning}:
    Here, each variable in the rule takes its value as a set of linguistic terms. I.e. if $X_i$ is a variable, and its term set is $\{l_1,l_2,l_3\}$ then in a rule variable $X_i$ can belong to the set  $\{l_1,l_2\}$. The variable can belong to the set of linguistic terms in a rule. This helps in reducing the number of rules to avoid the problem of increasing size. Here, a rule may be of the form:
    
    $\text{IF} \; x_1 \; is \; \{l_{11},\;l_{12}\} \; \text{and} \; x_2 \; is \; \{l_{23}\} \; \text{and} \; ... \; \text{and} \; X_n \; is \; \{l_{n1}, \; l_{n2}\} \; then \; y \; is \; Y$
    
    \item Takagi-Sugeno-Kang Fuzzy Rule-Based Systems:
    Takagi-Sugeno-Kang (TSK) \cite{takagi1985fuzzy, sugeno1988structure} FRBS considers a different form of rules, each rule in TSK-FRBS contains LVs as antecedents and a function of inputs as consequent. TSK FRBS models the output as a function of inputs and hence do not need a defuzzification process. This variant has been preferred in many applications where efficiency is of utmost importance. TSK-FRBS splits the input space into several fuzzy sub-spaces, based on the relationship between input and output. The main drawback of TSK-FRBS is its inability to provide interpretability for its input-output relationship. Here, the rule structure looks like:
    
    $\text{IF} \; x_1 \; is \; l_{1i} \; \text{and} \; x_2 \; is \; l_{2i} \; \text{and} \; ... \; \text{and} \; x_n \; is \; l_{ni} \; \text{then} \; y = p_0 + p_1*x_1 + ... + p_n*x_n$

    \item Approximate Mamdani Fuzzy Rule-based Systems:
    The DNF FRBS includes several items of term sets which can reduce the interpretability of the DNF FRBS. Approximate FRBSs \cite{duckstein1995fuzzy} are able to obtain better accuracy at the cost of loosing interpretability. Each rule in an approximate FRBS has its own fuzzy set instead of using linguistic terms. This approach generates semantic free rules, and has higher power of expressiveness due to the use of different fuzzy sets in each rule. It can adopt different number of rules depending on the complexity of the problem. In terms of drawbacks, approximate FRBSs suffers from the loss of interpretability, and also they can overfit the training data and perform poorly in case of unseen data. A typical rule here is of the form:

    \begin{figure}[!ht]
    \centering
    \includegraphics[width=0.35\textwidth]{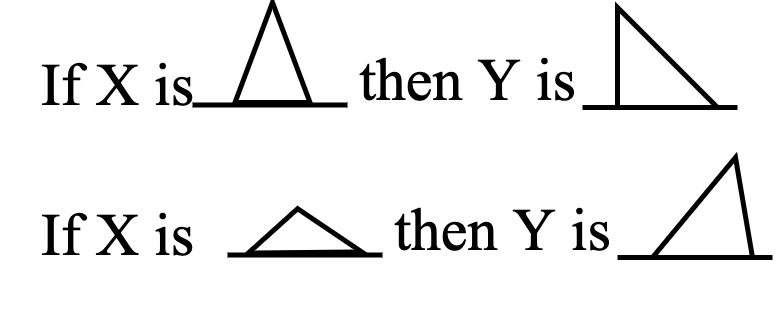}
    \label{Approximate mamdani}
\end{figure}
    
    \item Multiple Input Multiple Output (MIMO) FRBSs \cite{zeng1995approximation}:
    Here, as the name suggests, there are multiple outputs in a single rule rather than a single output variable in the previous variants of FRBSs. Outputs here are considered independently of each other and computed separately. The rule structure looks like:
    
    $\text{IF} \; x_1 \; is \; l_{1i} \; \text{and} \; x_2 \; is \; l_{2i} \; \text{and} \; ... \; \text{and} \; x_n \; is \; l_{ni} \; \text{then} \; y_1 \; is \; Y_1 \; \text{and} \; ... \; y_n \; is \; Y_n $
    
    \item Fuzzy Rule Based Classification Systems (FRBCSs):
    Fuzzy Rule Based Classification Systems (FRBCSs) \cite{chi1996fuzzy} is a system which uses fuzzy rules as a learning medium. In classical FRBSs, inputs are mapped to usually 1-D output but in FRBCSs inputs are mapped to one of the class labels. The rule structure looks like:
    
    $\text{IF} \; x_1 \; is \; l_{1i} \; \text{and} \; x_2 \; is \; l_{2i} \; \text{and} \; ... \; \text{and} \; x_n \; is \; l_{ni} \; then \; y = c$
\end{enumerate}

\section{Literature Review} \label{lit-review}

\begin{table}
    \centering
    \begin{adjustbox}{width=\textwidth}
        \begin{tabular}{l|l}
        \hline
        \hline
        Domain Focus & Keywords and Title Search\\
        \hline
        Fuzzy Systems & \begin{tabular} {@{}l@{}}Fuzzy Rule based Systems OR FRBS OR \\ Fuzzy System(s) OR FS OR Fuzzy Inference Systems OR FIS \end{tabular}\\
        \hline
        \begin{tabular} {@{}l@{}} Types of FRBSs \\ (GFS, NFS, HFS, eFS) \end{tabular} & \begin{tabular} {@{}l@{}} \textit{Type of FRBSs} OR \textit{Accronym} OR (\textit{Accronym} \\ OR \textit{Type of FRBSs} AND (Fuzzy Inference Systems OR FIS))\end{tabular}\\
        \hline
        \begin{tabular} {@{}l@{}l@{}} Types of FRBSs \\ (big data, interpretable, \\ imbalanced, clustering) \end{tabular} & \begin{tabular} {@{}l@{}l@{}} \textit{(Type of FRBSs AND (Fuzzy Systems OR FS))} \\ OR \textit{Accronym} OR (\textit{Accronym} OR \textit{Type of FRBSs} \\ AND (Fuzzy Inference Systems OR FIS))\end{tabular}\\
        \hline
        \hline
        \end{tabular}
    \end{adjustbox}
    \caption{Query structure to search for FRBSs and its type on Scopus database}
    \label{Query structure}
\end{table}

In this section, literature review and recent trends for Genetic/Evolutionary Fuzzy Systems \cite{herrera1996adaptation}, Hierarchical Fuzzy Systems \cite{gegov1995decomposition}, Adaptive Neuro Fuzzy Inference System \cite{jang1991fuzzy}, FRBSs for Big Data \cite{robles2009evolutionary}, FRBSs for Imbalanced Datasets \cite{batuwita2010fsvm}, Clusters centroids as rules in FRBSs \cite{yager1994generation} and FRBSs for Interpretability/Explainability \cite{hagras2018toward} are given. Each topic has one subsection which considers  articles in the Scopus database. Table \ref{Inclusion and Exclusion} highlights the inclusion and exclusion criteria for the papers to be considered in the paper and Table \ref{Query structure} shows the query structure used for Keywords or Title search in Scopus database. In Table \ref{Query structure}, \textit{Type of FRBSs} has been used as a proxy for different types of FRBSs such as Genetic fuzzy System; and \textit{Acronym} has been used as a proxy for the acronym of the type of FRBSs e.g. GFS/EFS for Genetic/Evolutionary Fuzzy System. The statistics related to each type of FRBSs are computed by finding the their name in the title or in the keywords. Areas which have seen a growing number of publication and citation have been considered for trends in each type of FRBSs. Table \ref{Advantages_disadvantages} summarizes the advantages, disadvantages, and the current trends in various types of fuzzy systems. 

\begin{table}
    \centering
    \begin{adjustbox}{width=\textwidth}
        \begin{tabular}{ll}
        \hline
        \hline
            Inclusion Criteria & \begin{tabular} {@{}c@{}c@{}c@{}}Journal papers, book chapters, articles and conference proceedings\\ Written in English \\ Published during 2010-2021 \\ Affiliated to FRBSs \end{tabular} \\
            \hline
            Exclusion Criteria & \begin{tabular} {@{}c@{}c@{}c@{}c@{}} Articles not concerning Fuzzy rule-based System or Fuzzy Systems\\ Unpublished articles \\ Duplicate articles \\ Articles published in language other than English \\ Tutorials, blogs and other non-scientific work \end{tabular}  \\
        \hline
        \hline  
        \end{tabular}
    \end{adjustbox}
    \caption{Inclusion and Exclusion criteria for literature review in Scopus database}
    \label{Inclusion and Exclusion}
\end{table}

\subsection{Genetic/Evolutionary Fuzzy Systems:}
A GFS is a type of Fuzzy Rule Based System which employs evolutionary algorithms \cite{back1993overview} such as genetic algorithm \cite{holland1992adaptation}, and genetic programming \cite{koza1994genetic} for learning purpose in the fuzzy system. Designing FRBSs can be seen as a search problem, e.g., finding the set of rules, tuning the membership parameters. GAs (EAs) are known to handle the large search space for near optimal solution with a performance measure. Apart from the ability to explore a large search space, GAs can incorporate prior knowledge, in case of fuzzy systems, prior knowledge can be the number of rules, membership functions, linguistic variables and so on.

\begin{figure}
    \centering
    \includegraphics[width=\textwidth]{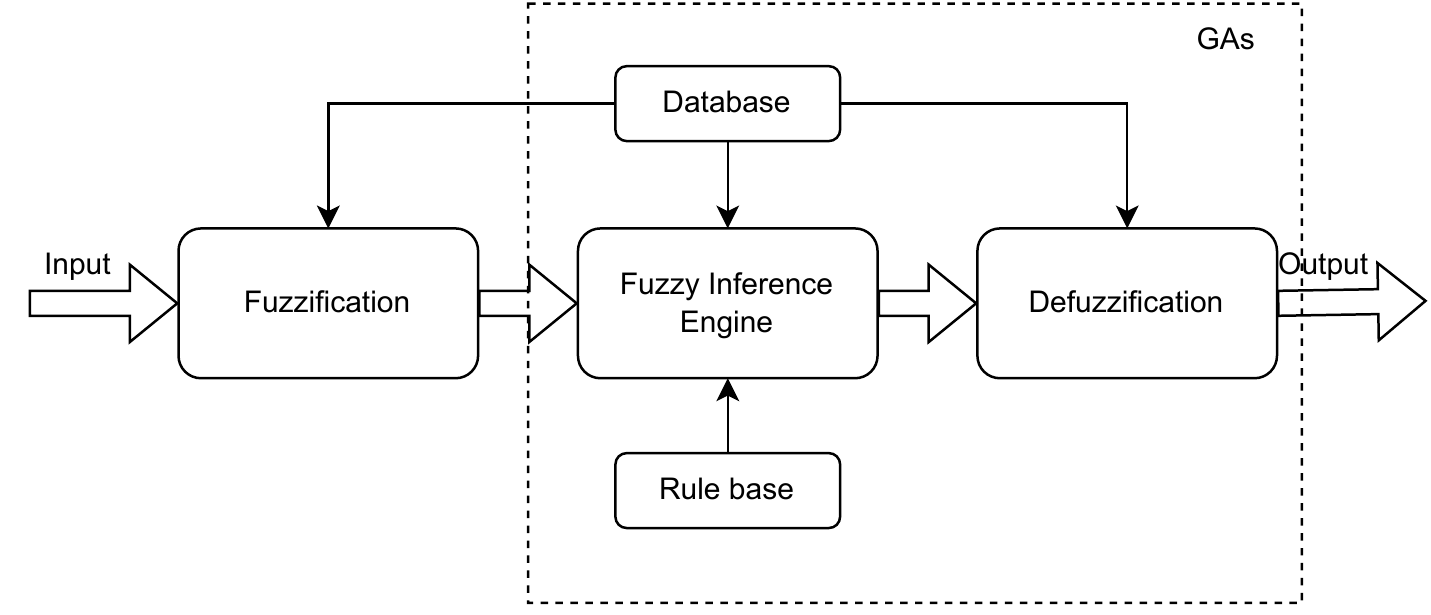}
    \caption{Structure of Genetic Fuzzy System (GFS)}
    \label{GFS_generic}
\end{figure}

In GFS, the first step is to identify the areas of FRBSs which can be optimized using GAs or EAs. Fig. \ref{GFS_generic} shows the areas in which GAs can be used for learning. In general, learning in FRBSs is of two types: learning the KB components and tuning the parameters. Learning KB components can be further divided into four categories namely, rule learning, rule selection, learning DB components, and learning KB simultaneously. Rule learning approaches learns the rules automatically from the data for a predefined DB. Thrift et. al \cite{thrift1991fuzzy} presents the classical proposal on tuning. The generated number of rules can be huge, rule selection selects the best set of rules, \cite{ishibuchi1995selecting} presents the first contribution in the field. Learning of DB components involves learning shape of membership functions, number of fuzzy sets and other DB components. Cordon et al. \cite{cordon2001generating} is the pioneer paper in the field. Learning KB components involves learning DB components and rules simultaneously, this leads to a large search space which makes learning more difficult, first such methodology was presented in \cite{homaifar1995simultaneous}. Tuning of the parameters is done to improve the system’s performance. Tuning can be done by adjusting the membership function parameters, by adjusting inference systems parameters and by adjusting weights in defuzzification methods. GFS suffers from the disadvantage of genetic algorithms which is computational complexity I.e. GAs search for a solution may take significant time. In the literature, optimized exploration and exploitation can improve the running time of GAs. Examples include an arithmetic optimization algorithm that does not need to adjust many parameters \cite{abualigah2021arithmetic}, pairie dog optimization \cite{ezugwu2022prairie}, \cite{abualigah2022reptile} that uses meta-heuristic animal based activities to optimize exploration and exploitation.

In the years 2010-2021, GFSs literature saw several important contributions such as in Hadavandi et al. \cite{hadavandi2010integration}) have used GFS with Self-organizing maps clustering for next day forecasting. Elhag et al. \cite{elhag2015combination} introduced a pairwise learning framework with GFS for intrusion detection. GM3M index proposed by Gacto et al. \cite{gacto2010integration} is a geometric mean for three metrics accuracy, maximizing semantic interpretability, minimizing rule complexity. The authors use SPEA2 MOEA (Multi-objective Evolutionary Algorithm) to generate pareto optimal solutions for the three metrics. Methodology presented by Alcala et al. \cite{alcala2011fast} deals with the problem of regression in high dimensional data, they also focus on developing simpler rules by adjusting linguistic fuzzy partitions, and in Sanz et al. \cite{sanz2011genetic} interval-valued fuzzy sets were used to model the linguistic labels, amplitude for IVFSs was tuned using weak ignorance theorem. 

There exists very few review papers in the literature for genetic fuzzy systems. First review paper by Cordon et al. \cite{cordon2001ten} talked about the GFS models, its application, trends, and open questions. A short overview of GFS models was also presented in Herrera et al. \cite{herrera2005genetic}. In \cite{herrera2008genetic}, Herrera reviewed the taxonomy, GFS models, advances in the field of GFS, trends and the future areas related to GFSs. A review for learning Mamdani-type fuzzy rule-based system was provided in Cordon et al. \cite{cordon2011historical}. The paper focused on tools for improving accuracy while designing interpretable GFSs, \cite{cordon2011historical} focuses on aspects other than accuracy (e.g. interpretability), and on reducing the complexity of the model. Synthesis of eFS with special focus on Genetic programming based eFS was given in Koshiyama et al. \cite{koshiyama2019automatic}. Fernandez et al. \cite{fernandez2019evolutionary} have presented an overview of Evolutionary fuzzy Systems, their terminologies, applications, areas where evolutionary approaches are useful and knowledge base generation.

\begin{table}
    \centering
    \begin{adjustbox}{width=\textwidth}
        \begin{tabular}{l|l|l|l|l|l|l|l|l|l|l}
        \hline
        \hline
            number & Journal Name & Publisher & GFS & HFS & NFS & eFS & Big data & imbalanced & interpretabe-FS & clustering \\
            \hline
            1 &	\begin{tabular} {@{}c@{}}IEEE International Confer- \\ ence of Fuzzy Systems \end{tabular}& IEEE & 46 & 16 & 17 & 5 & 9 & 13 & 101 & 5 \\
            \hline
            2 &	\begin{tabular} {@{}c@{}} Advances in Intelligent \\ System and computing \end{tabular}& Springer & 24 & 6 & 43 & 3 & 3 & 2 & 15 & 15 \\
            \hline
            3 &	Applied Soft Computing & Elsevier & 23 & 4 & 29 & 11 & 1 & 2 & 16 & 2 \\
            \hline
            4 &	\begin{tabular} {@{}c@{}} Expert Systems with \\ Application \end{tabular} & Elsevier & 21 & 2 & 44 & 4 & - & 1 & 13 & 8 \\
            \hline
            5 &	\begin{tabular} {@{}c@{}} IEEE Transactions on \\ Fuzzy Sets and Systems \end{tabular} & IEEE & 12 & 8 & 10 & 15 & 2 & 3 & 35 & 4 \\
            \hline
            6 &	Soft Computing & Springer & 11 & 4 & 22 & 1 & - & - & 12 & 6 \\
            \hline
            7 &	\begin{tabular} {@{}c@{}} International journal of \\ fuzzy systems \end{tabular} & Springer & 5 & 2 & 12 & - & - & 2 & 3 & 2 \\
            \hline
            8 &	Information Sciences & Elsevier & 8 & 2 & 3 & 8 & 1 & 6 & 18 & 2 \\
            \hline
            9 & IEEE Access & IEEE & 3 & 4 & 20 & 3 & 2 & 2 & 8 & 3 \\
            \hline
            10 & Fuzzy sets \& Systems & Elsevier & 17 & 5 & 7 & 3 & 3 & 2 & 9 & 2 \\
            \hline
            11 & Knowledge Based Systems & Elsevier & 7 & 2 & 5 & 1 & 1 & 4 & 6 & - \\
        \hline
        \hline
        \end{tabular}
    \end{adjustbox}
    \caption{List of top journals/conferences with the number of articles published in various types of FRBSs between the years 2010-2021}
    \label{FS_publishers}
\end{table}

\begin{figure}
\centering
    \begin{subfigure}[b]{0.45\textwidth}
         \centering
         \includegraphics[width=\textwidth]{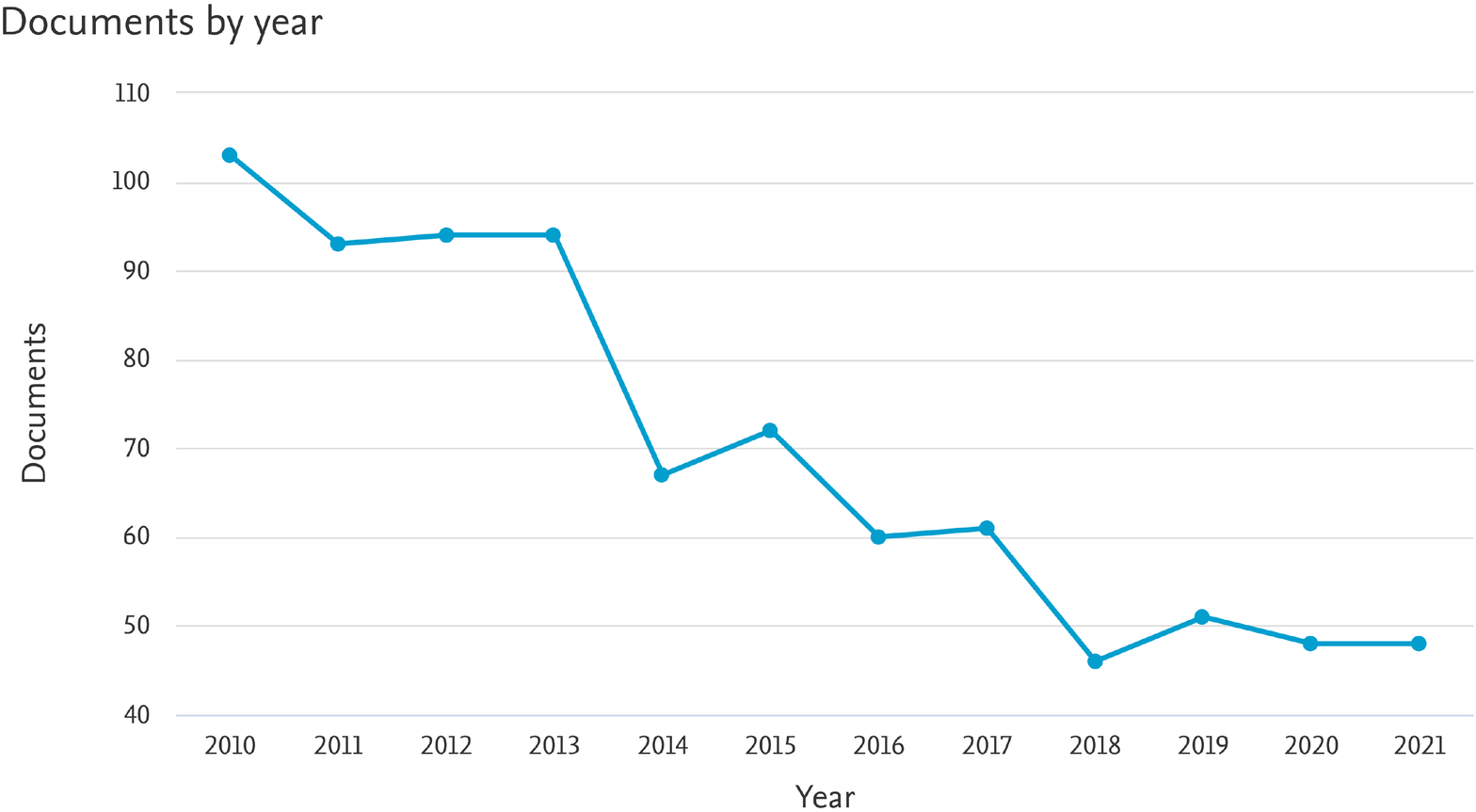}
         \caption{ }
         \label{GFS publications 2010-2021}
     \end{subfigure}
     \hfill
     \begin{subfigure}[b]{0.45\textwidth}
         \centering
         \includegraphics[width=\textwidth]{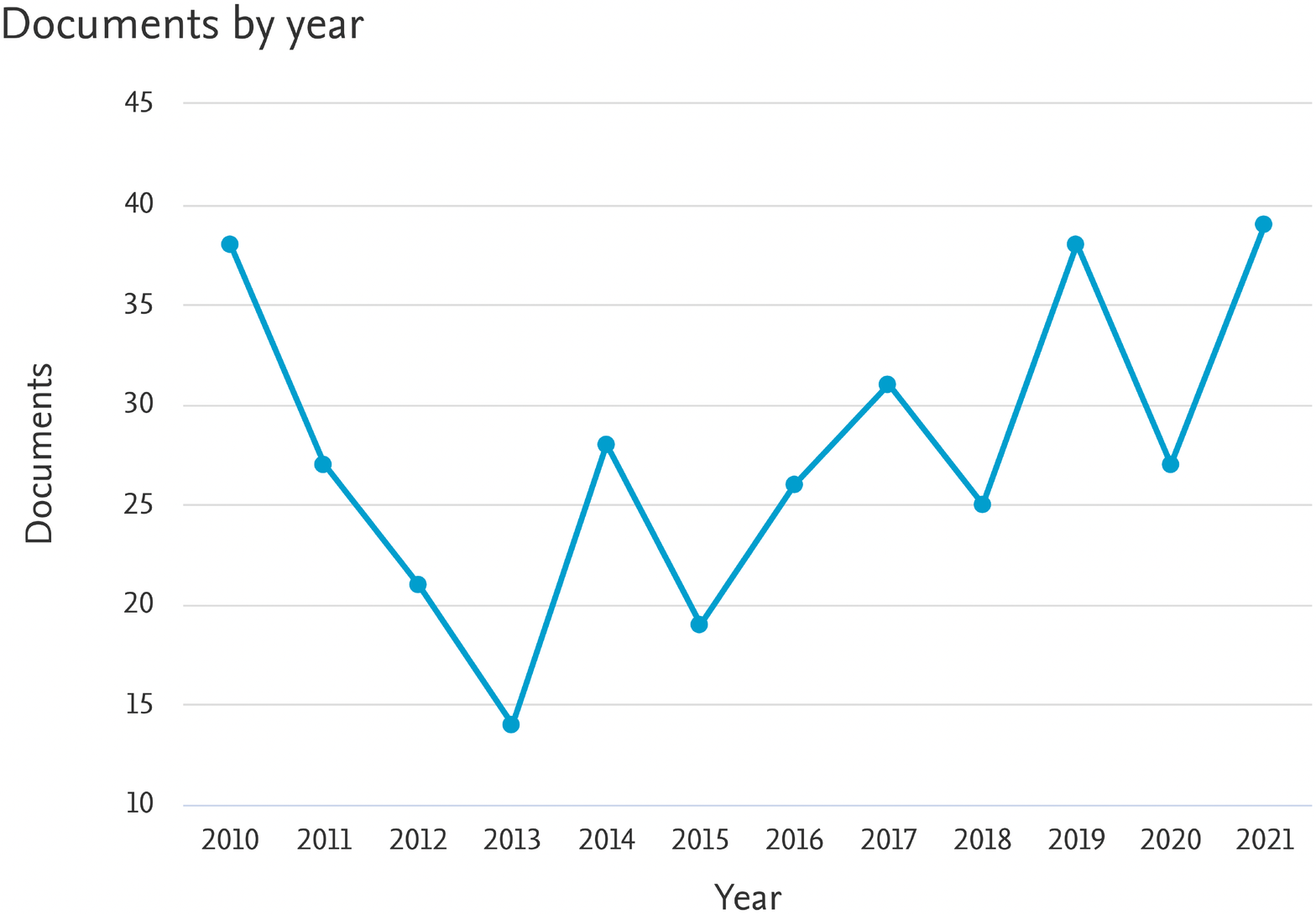}
         \caption{ }
         \label{HFS publications 2010-2021}
     \end{subfigure}
     \hfill
     \begin{subfigure}[b]{0.45\textwidth}
         \centering
         \includegraphics[width=\textwidth]{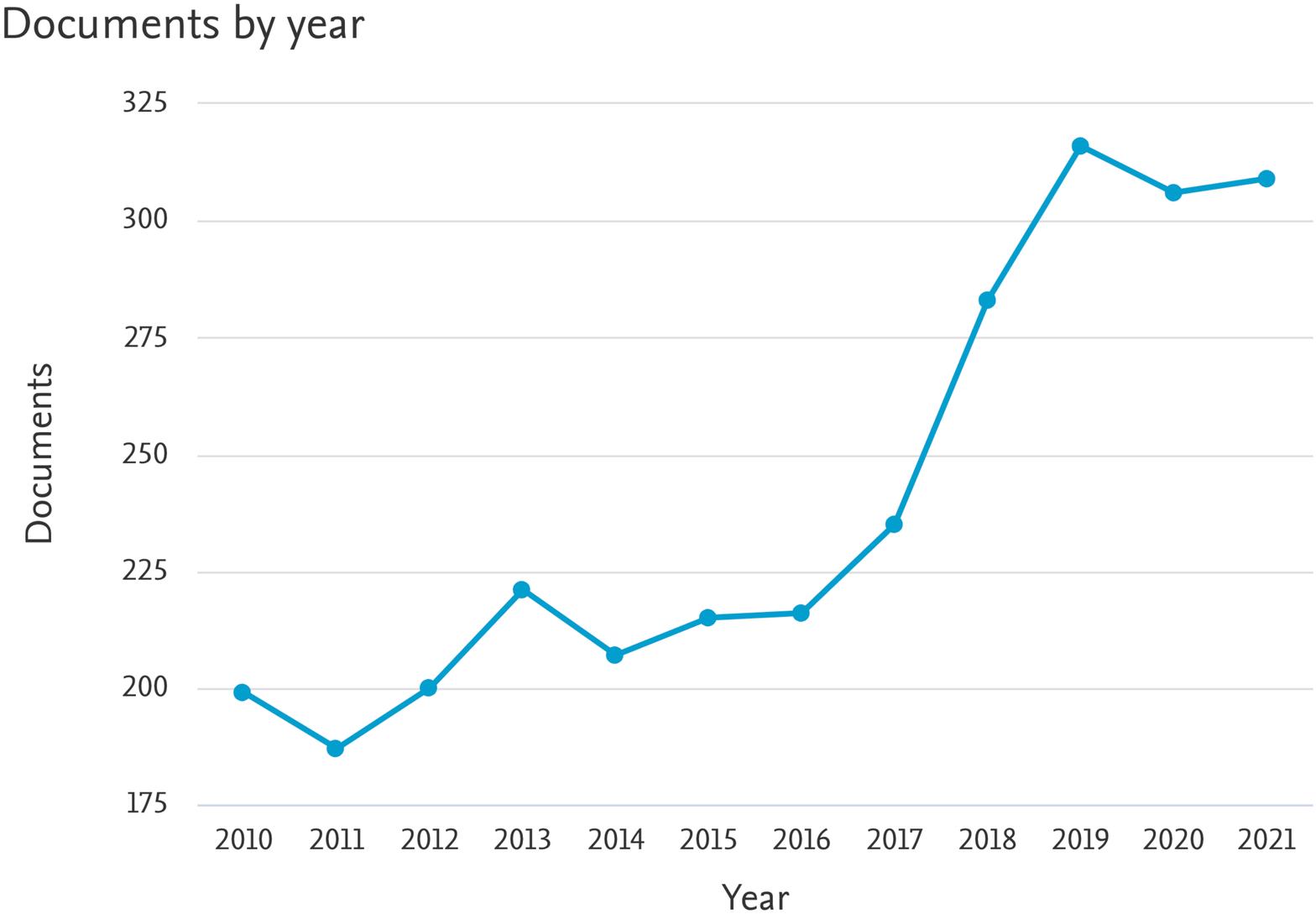}
         \caption{ }
         \label{NFS publications 2010-2021}
     \end{subfigure}
     \hfill
     \begin{subfigure}[b]{0.45\textwidth}
         \centering
         \includegraphics[width=\textwidth]{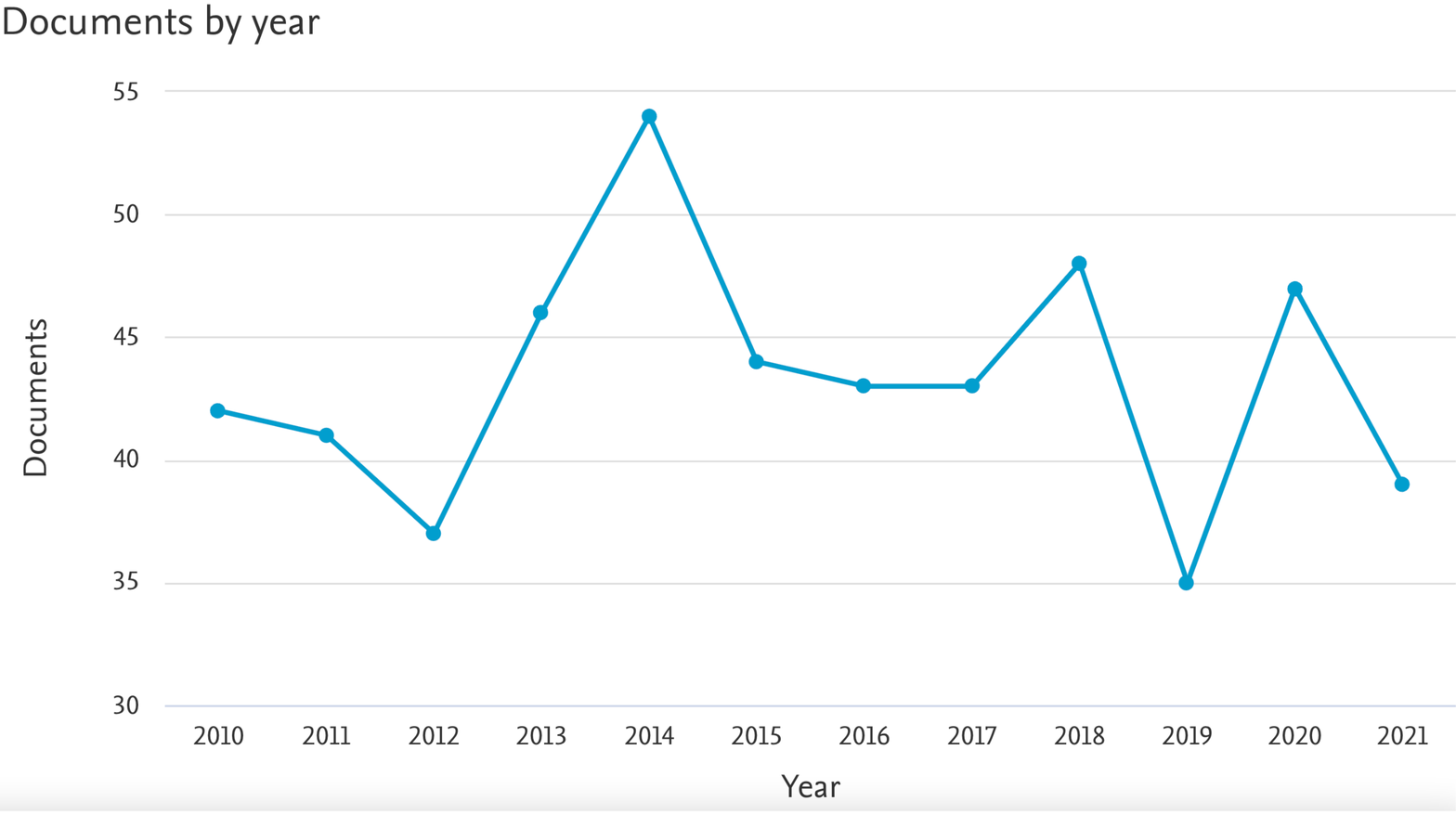}
         \caption{ }
         \label{eFS publications 2010-2021}
     \end{subfigure}
     \hfill
     \begin{subfigure}[b]{0.45\textwidth}
         \centering
         \includegraphics[width=\textwidth]{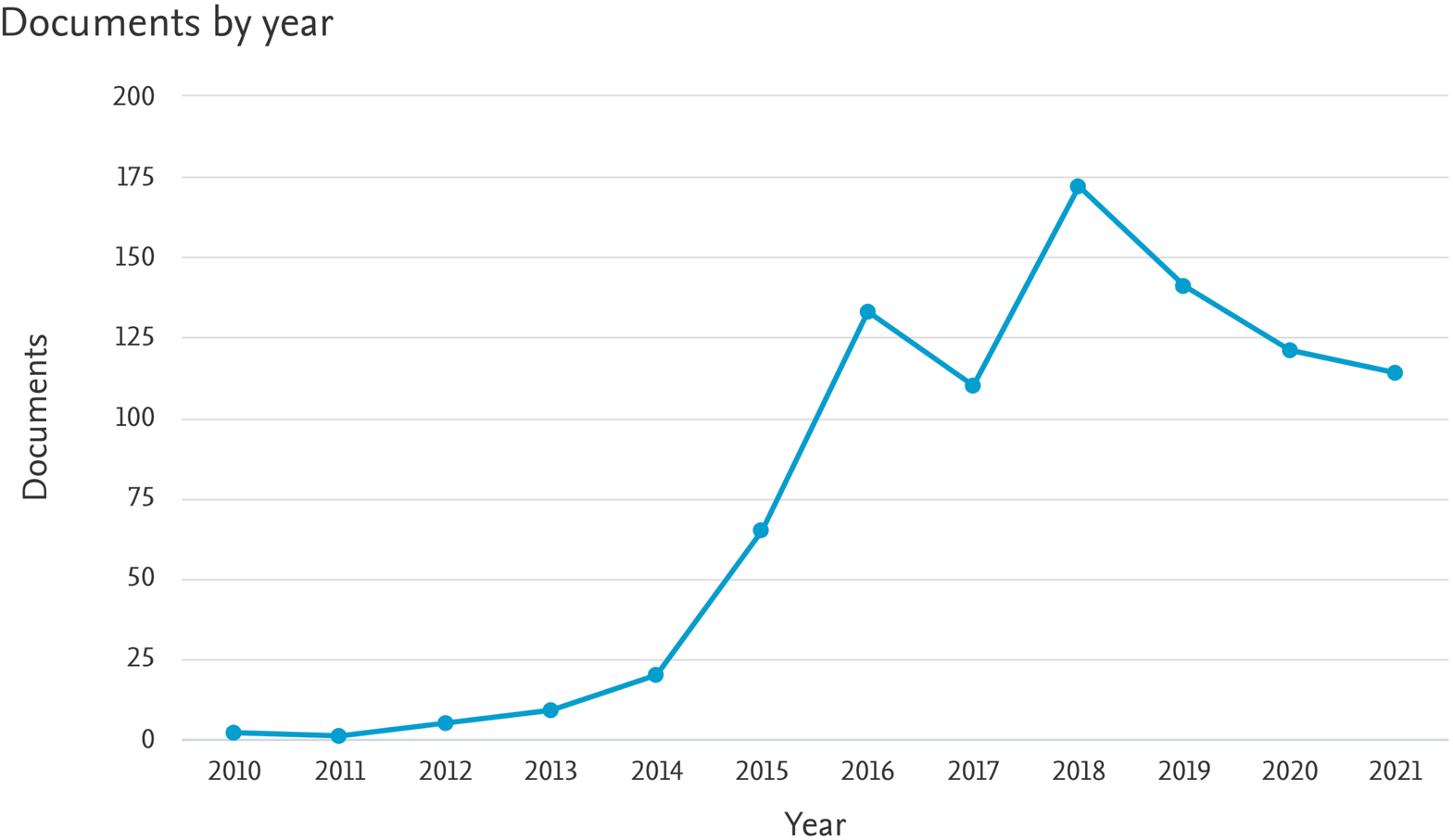}
         \caption{ }
         \label{Big data publications 2010-2021}
     \end{subfigure}
     \hfill
     \begin{subfigure}[b]{0.45\textwidth}
         \centering
         \includegraphics[width=\textwidth]{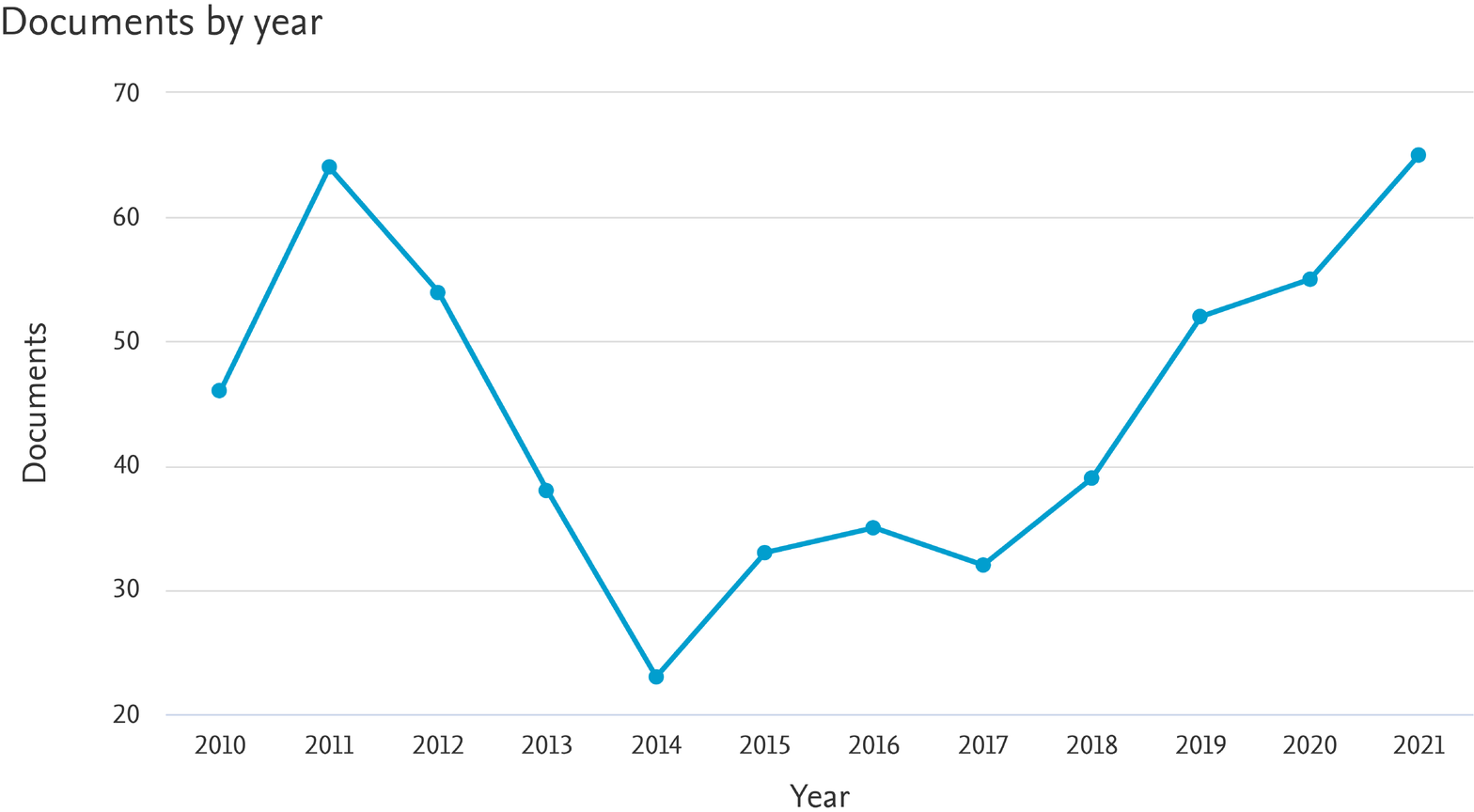}
         \caption{ }
         \label{XAI publications 2010-2021}
     \end{subfigure}
     \begin{subfigure}[b]{0.45\textwidth}
         \centering
         \includegraphics[width=\textwidth]{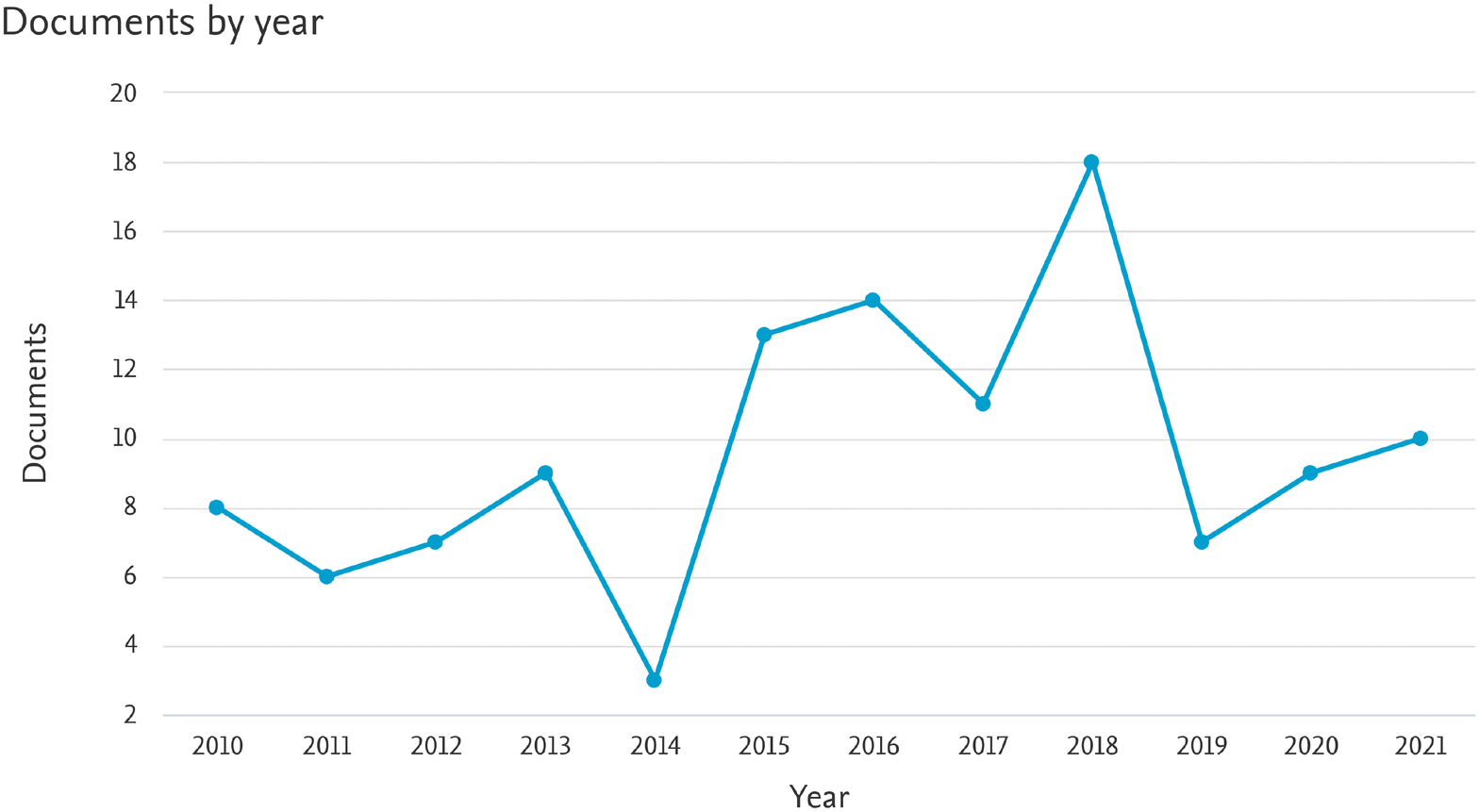}
         \caption{ }
         \label{imbalance publications 2010-2021}
     \end{subfigure}
     \hfill
     \begin{subfigure}[b]{0.45\textwidth}
         \centering
         \includegraphics[width=\textwidth]{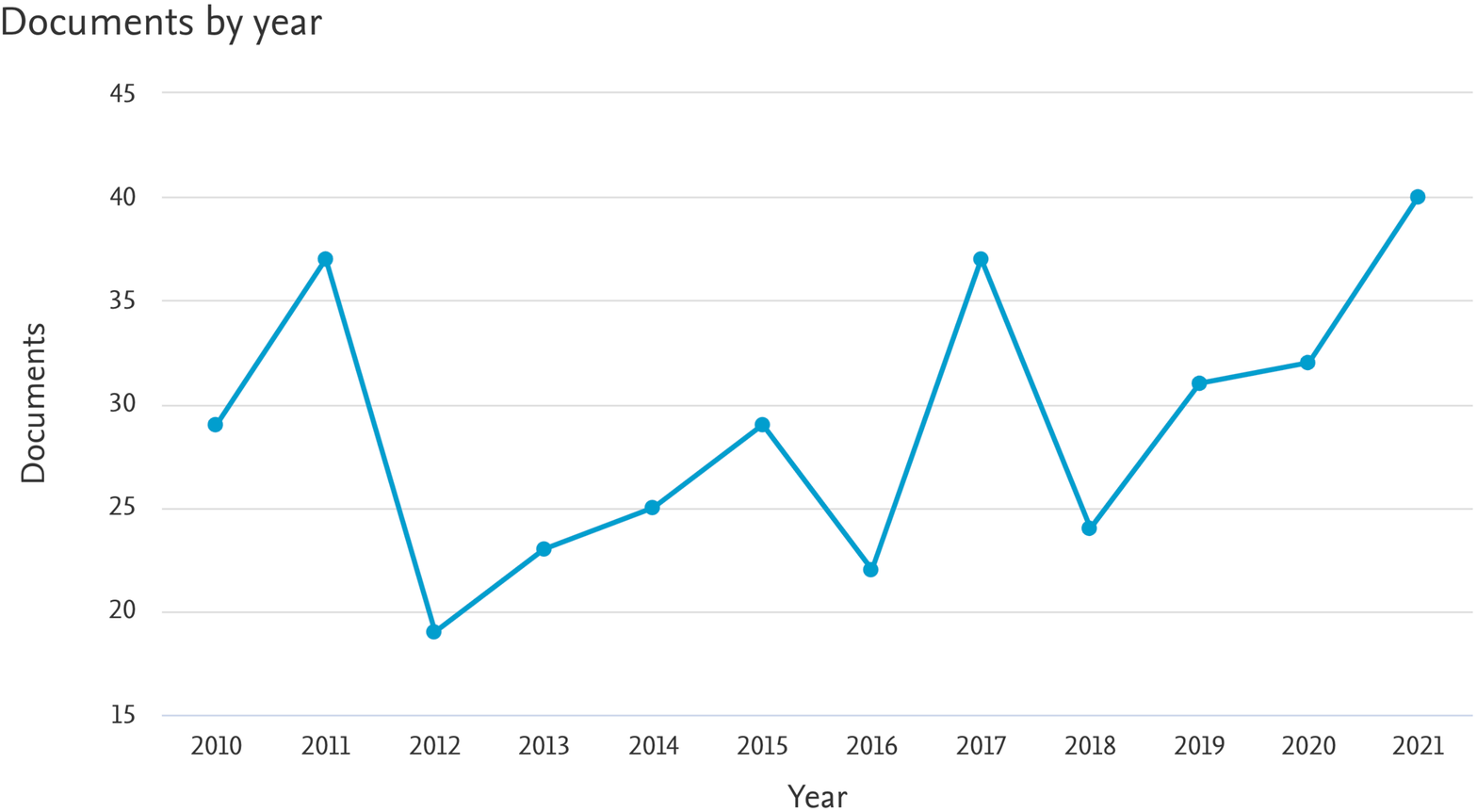}
         \caption{ }
         \label{cluster publications 2010-2021}
     \end{subfigure}
\caption{ Number of published articles during the year 2010-2021: (a) Genetic Fuzzy Systems, (b) Hierarchical Fuzzy Systems, (c) Neuro-Fuzzy Systems, (d) Evolving Fuzzy Systems, (e) FRBSs for Big Data, (f) Interpretability/Explainability in FRBSs, (g) FRBSs for imbalanced data, (h) Cluster centroids as rules in FRBSs.}
\label{No pubs 2010-2021}
\end{figure}

\begin{figure}
\centering
    \begin{subfigure}[b]{0.45\textwidth}
         \centering
         \includegraphics[width=\textwidth]{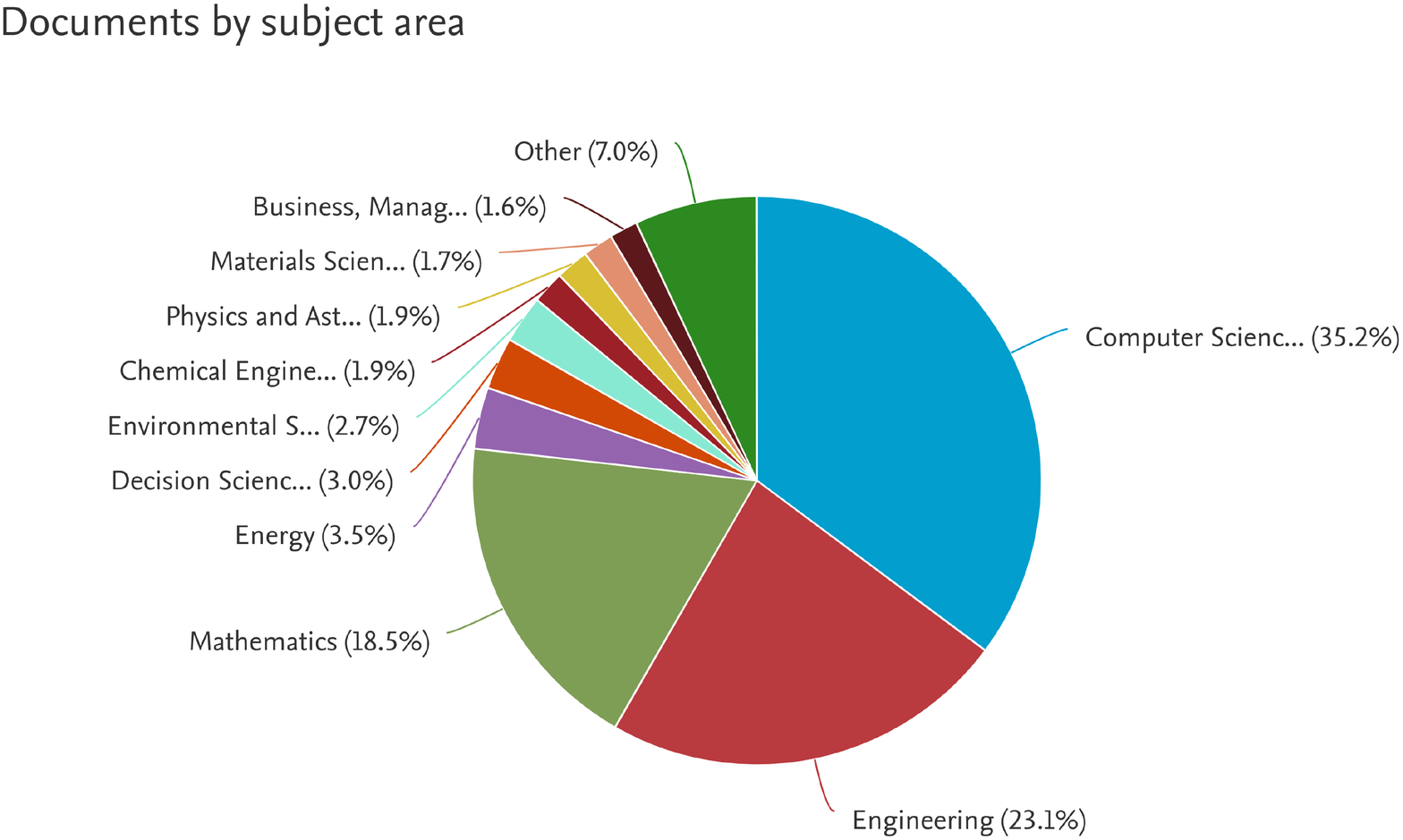}
         \caption{ }
         \label{Area-wise publications in GFSs}
     \end{subfigure}
     \hfill
    \begin{subfigure}[b]{0.45\textwidth}
         \centering
         \includegraphics[width=\textwidth]{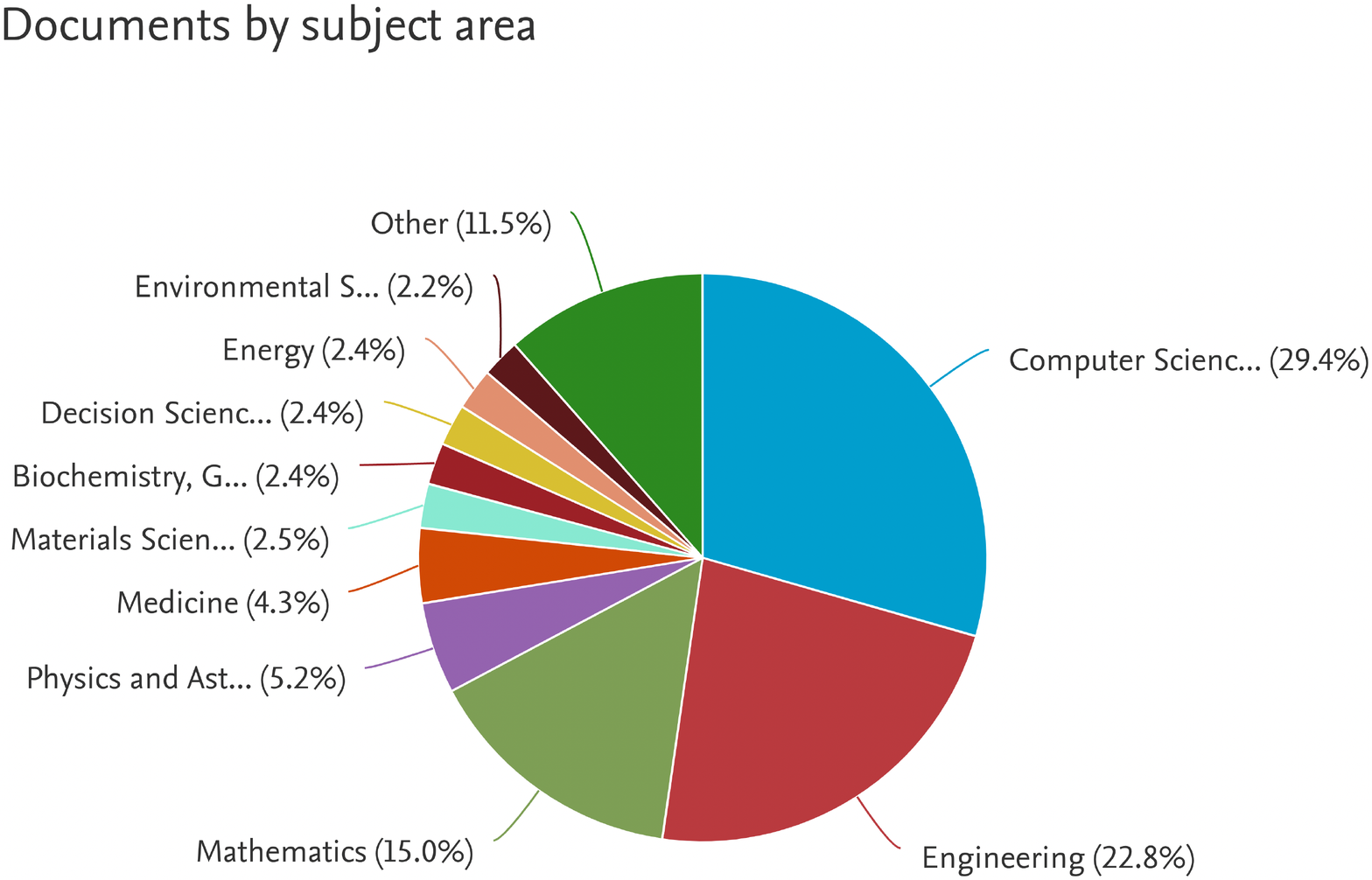}
         \caption{ }
         \label{Area-wise publications in HFSs}
     \end{subfigure}
     \hfill
     \begin{subfigure}[b]{0.45\textwidth}
         \centering
         \includegraphics[width=\textwidth]{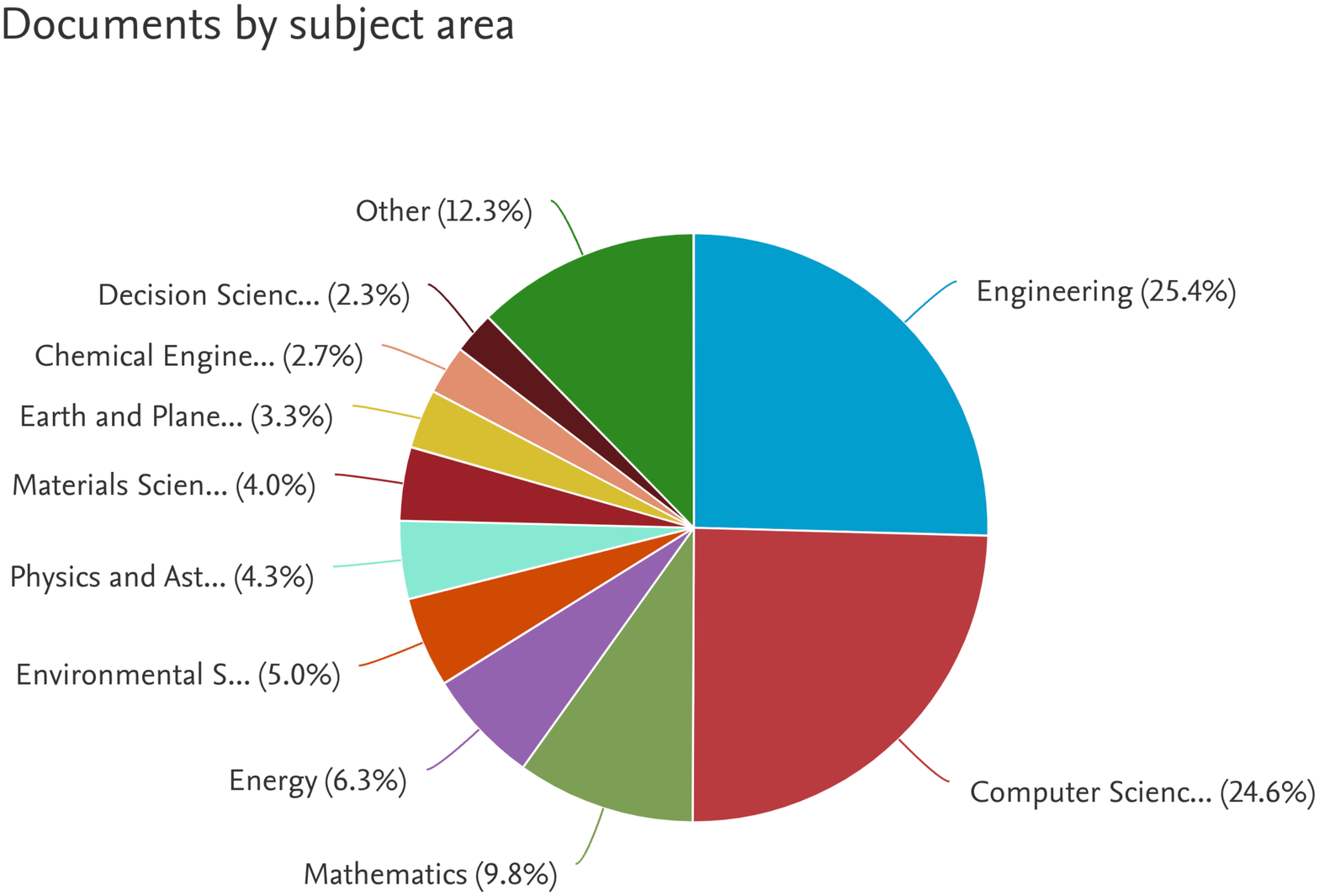}
         \caption{ }
         \label{Area-wise publications in NFSs}
     \end{subfigure}
     \hfill
     \begin{subfigure}[b]{0.45\textwidth}
         \centering
         \includegraphics[width=\textwidth]{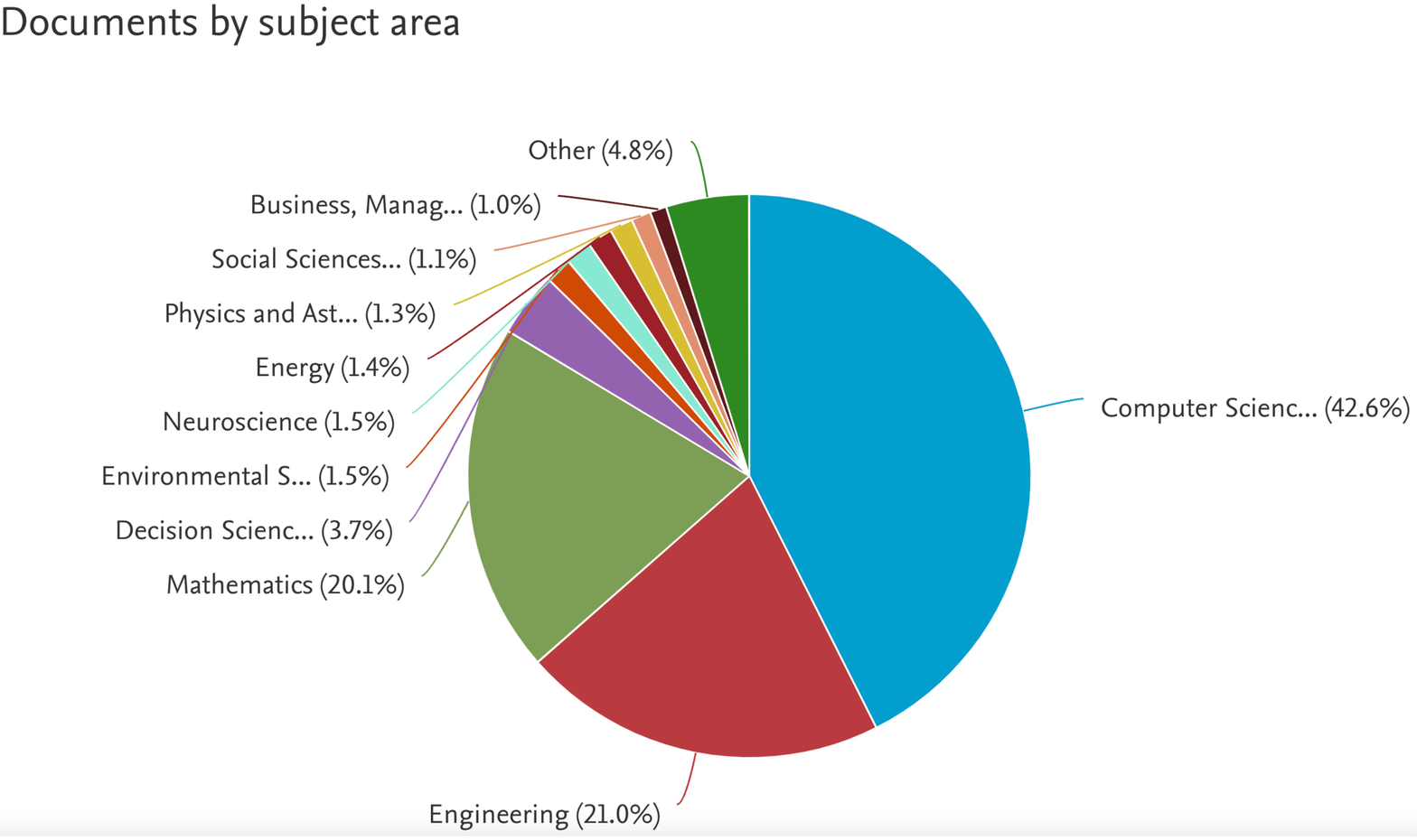}
         \caption{ }
         \label{Area-wise publications in eFSs}
     \end{subfigure}
     \hfill
     \begin{subfigure}[b]{0.45\textwidth}
         \centering
         \includegraphics[width=\textwidth]{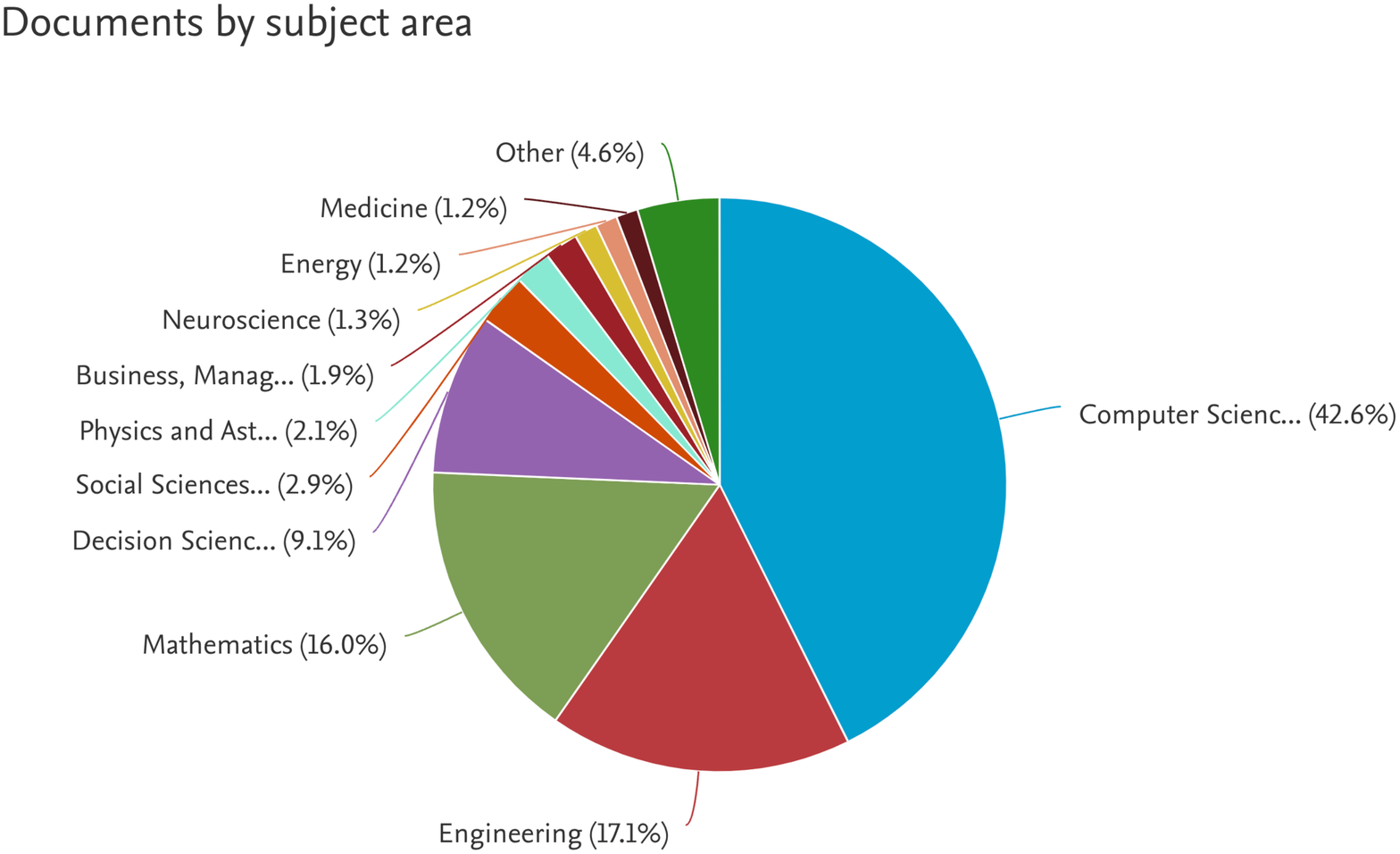}
         \caption{ }
         \label{Area-wise publications Big Data}
     \end{subfigure}
     \hfill
     \begin{subfigure}[b]{0.45\textwidth}
         \centering
         \includegraphics[width=\textwidth]{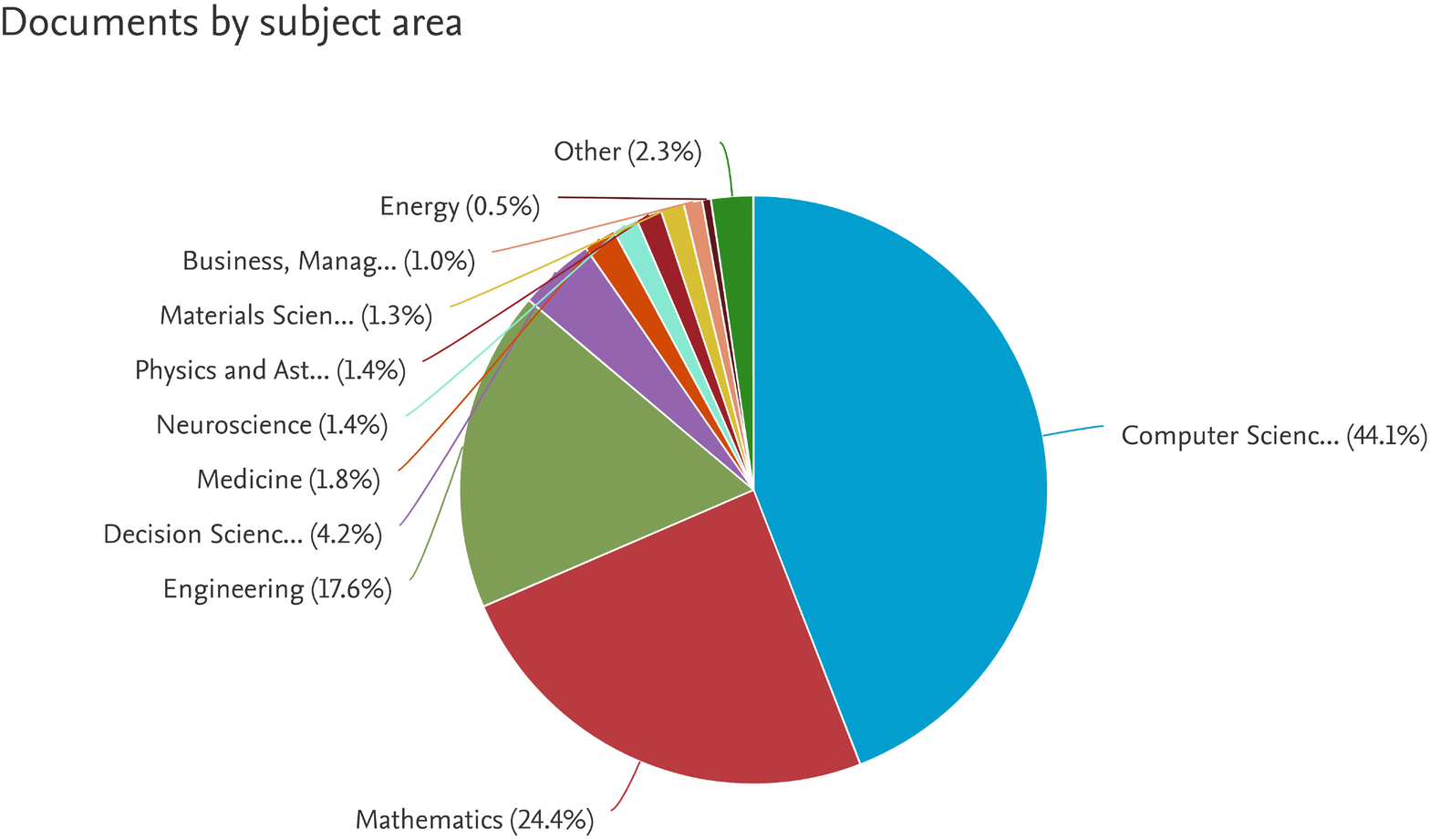}
         \caption{ }
         \label{Area-wise publications XAI}
     \end{subfigure}
\caption{ Area-wise categorization for the published articles in FRBSs during 2010-2021: a) Genetic Fuzzy Systems, (b) Hierarchical Fuzzy Systems, (c) Neuro-Fuzzy Systems, (d) Evolving Fuzzy Systems, (e) FRBSs for Big Data, (f) Interpretability/Explainability in FRBSs, (g) FRBSs for imbalanced data, (h) Cluster centroids as rules in FRBSs.}
\label{Area-wise 2010-2021}
\end{figure}

\begin{figure}
\centering
    \begin{subfigure}[b]{0.45\textwidth}
         \centering
         \includegraphics[width=\textwidth]{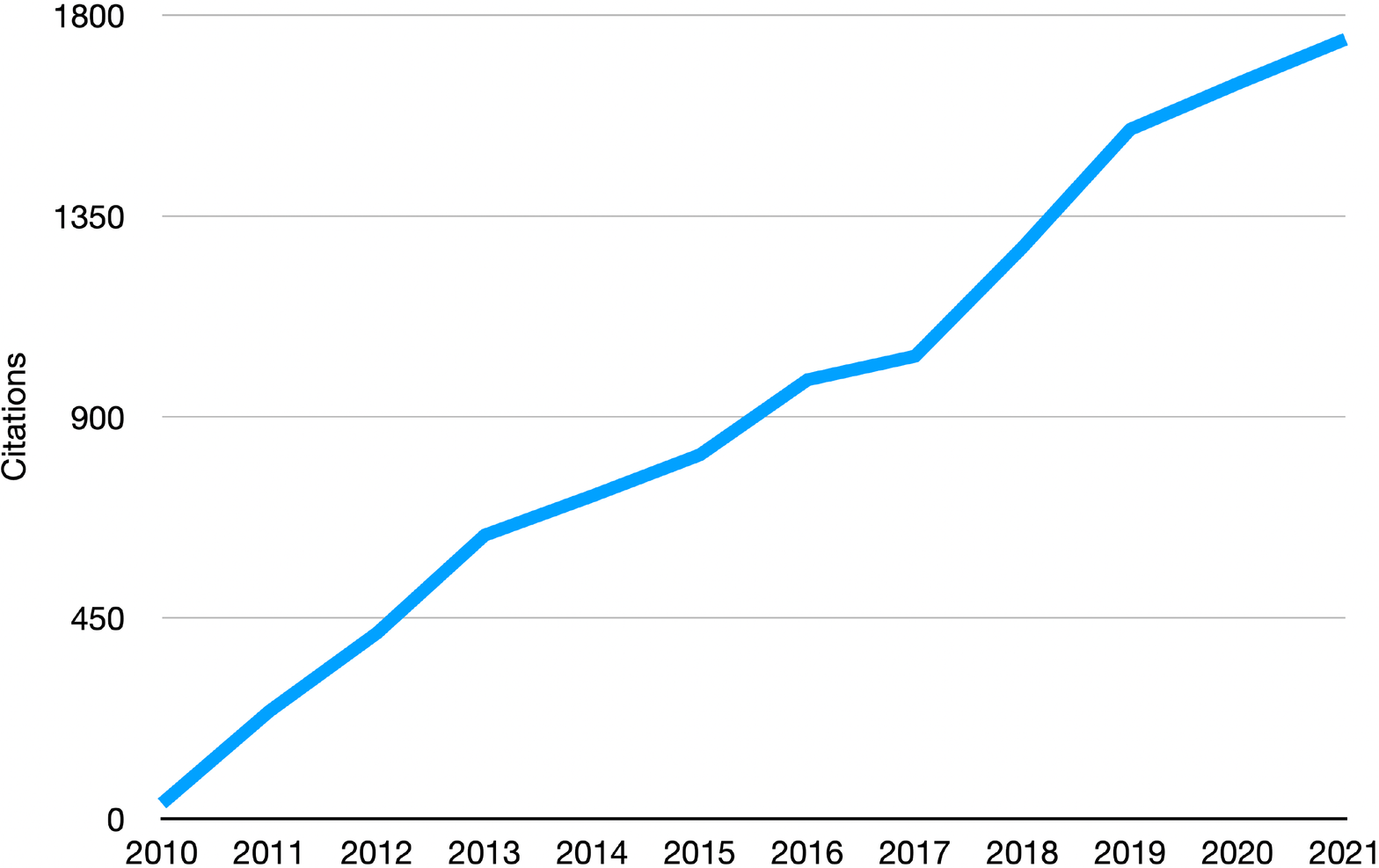}
         \caption{ }
         \label{Citations in the field of GFSs 2010-2021}
     \end{subfigure}
     \hfill
     \begin{subfigure}[b]{0.45\textwidth}
         \centering
         \includegraphics[width=\textwidth]{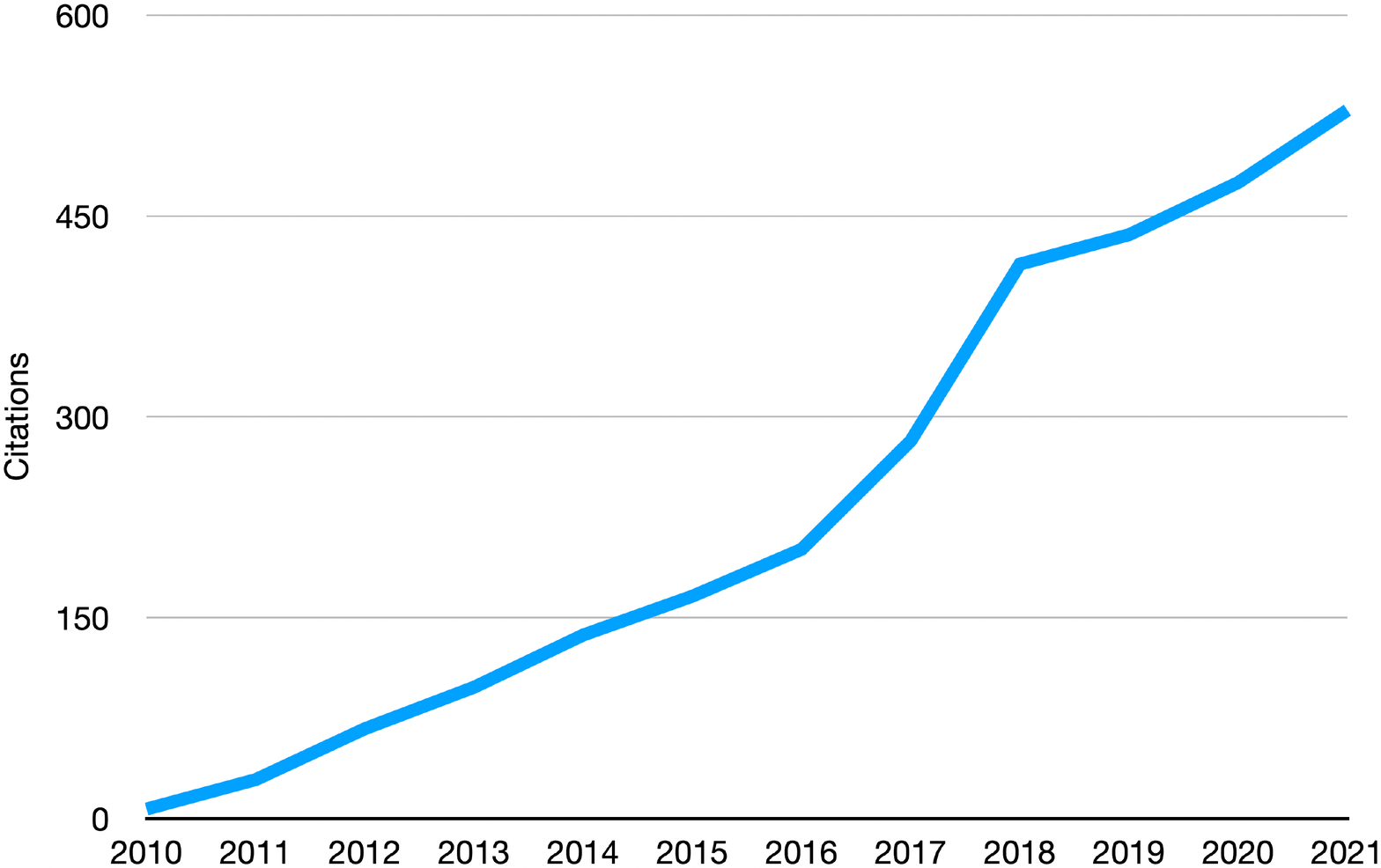}
         \caption{ }
         \label{Citations in the field of HFSs 2010-2021}
     \end{subfigure}
     \hfill
     \begin{subfigure}[b]{0.45\textwidth}
         \centering
         \includegraphics[width=\textwidth]{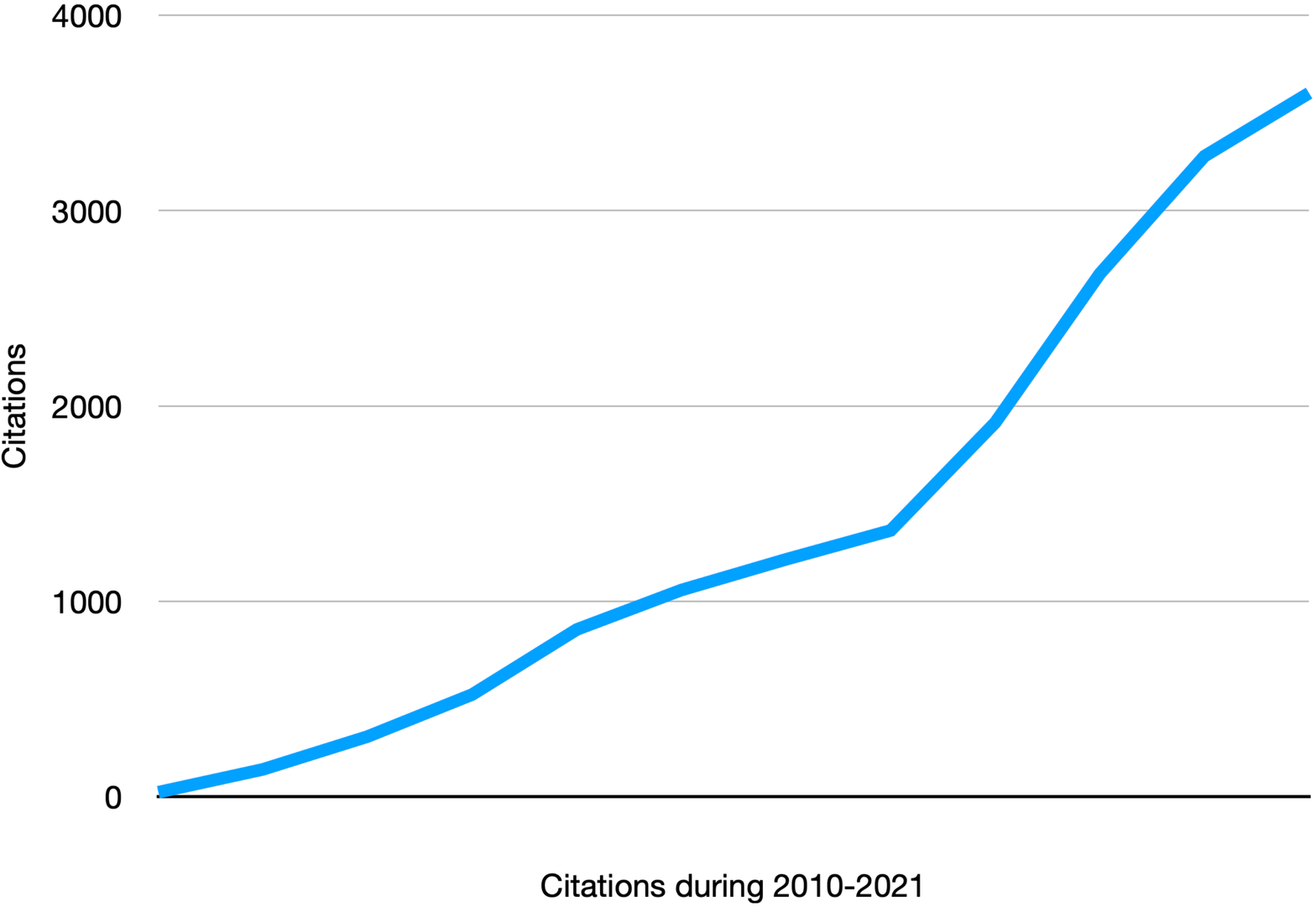}
         \caption{ }
         \label{Citations in the field of NFSs 2010-2021}
     \end{subfigure}
     \hfill
     \begin{subfigure}[b]{0.45\textwidth}
         \centering
         \includegraphics[width=\textwidth]{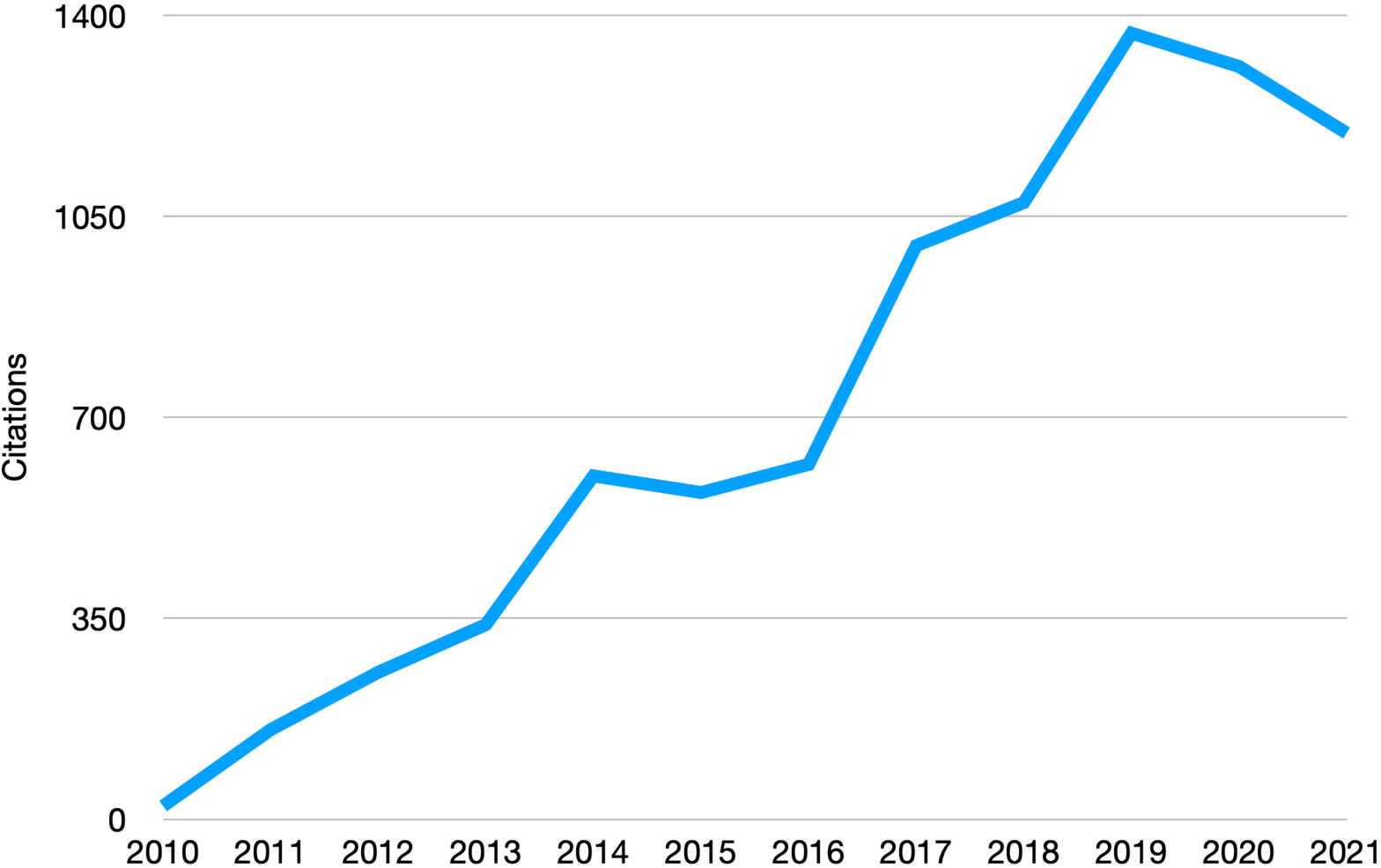}
         \caption{ }
         \label{Citations in the field of eFSs 2010-2021}
     \end{subfigure}
     \hfill
     \begin{subfigure}[b]{0.45\textwidth}
         \centering
         \includegraphics[width=\textwidth]{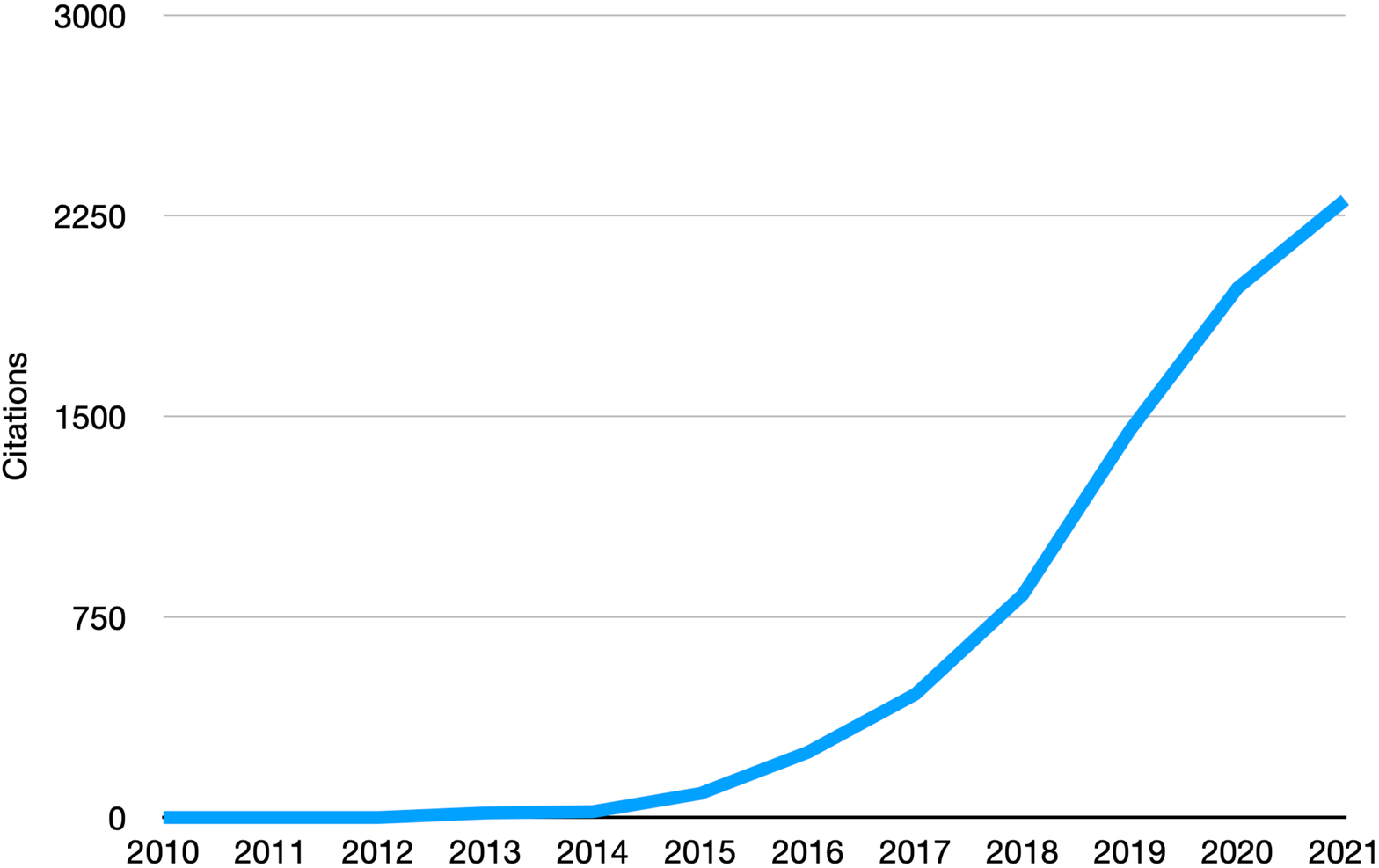}
         \caption{ }
         \label{Citations in the field of Big Data 2010-2021}
     \end{subfigure}
     \hfill
     \begin{subfigure}[b]{0.45\textwidth}
         \centering
         \includegraphics[width=\textwidth]{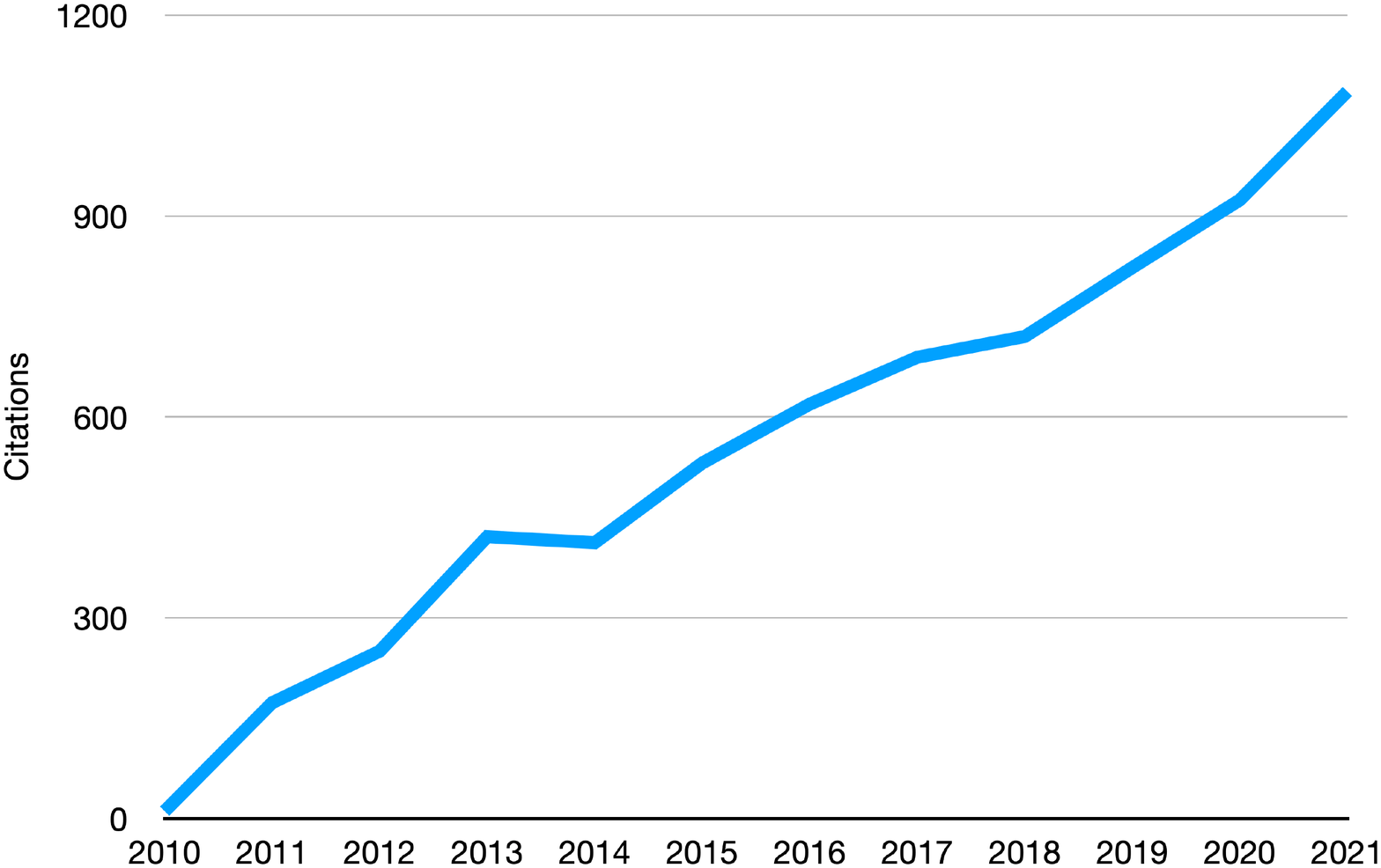}
         \caption{ }
         \label{Citations in the field of XAI 2010-2021}
     \end{subfigure}
     \hfill
     \begin{subfigure}[b]{0.45\textwidth}
         \centering
         \includegraphics[width=\textwidth]{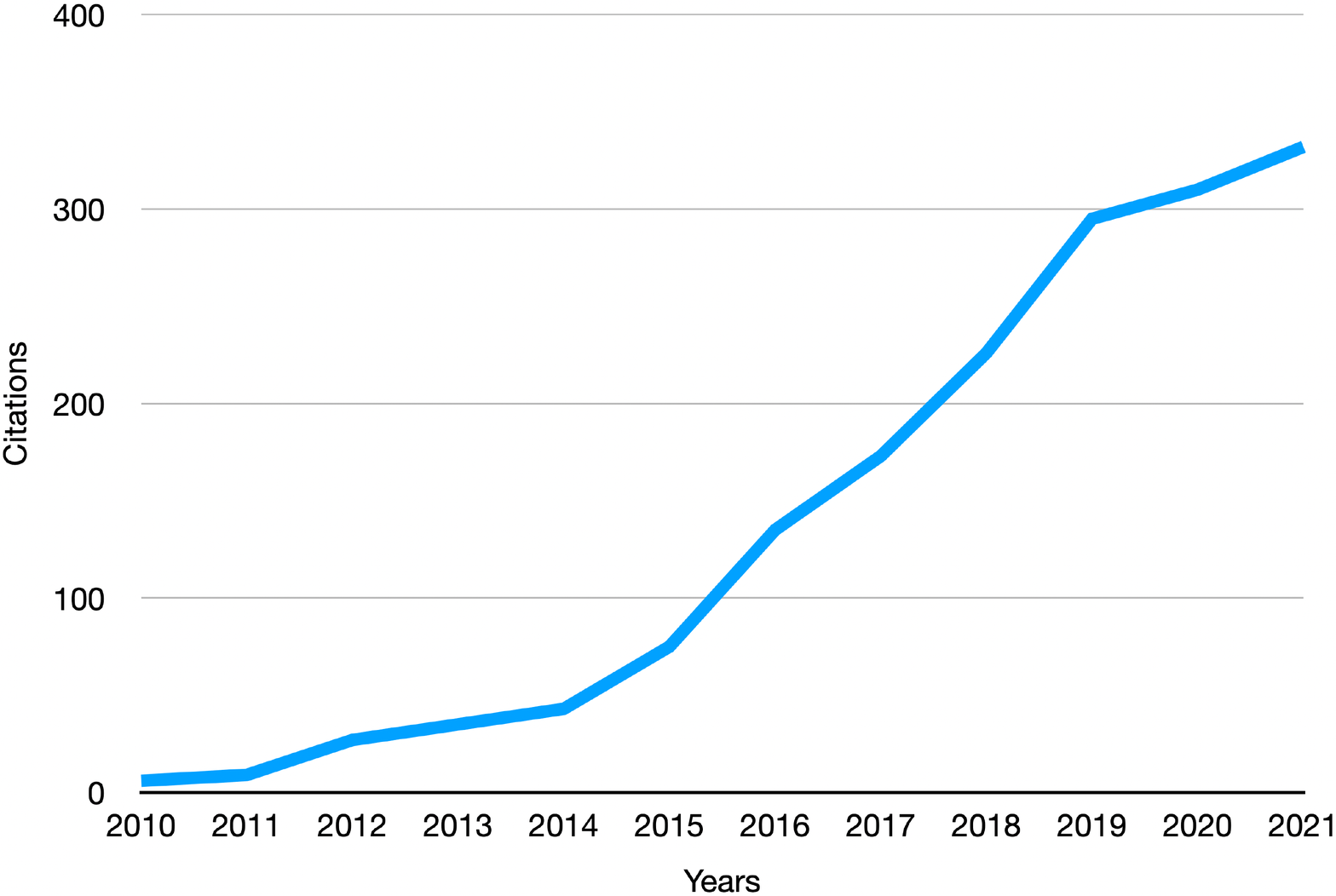}
         \caption{ }
         \label{cites imbalanced FRBS}
     \end{subfigure}
     \hfill
     \begin{subfigure}[b]{0.45\textwidth}
         \centering
         \includegraphics[width=\textwidth]{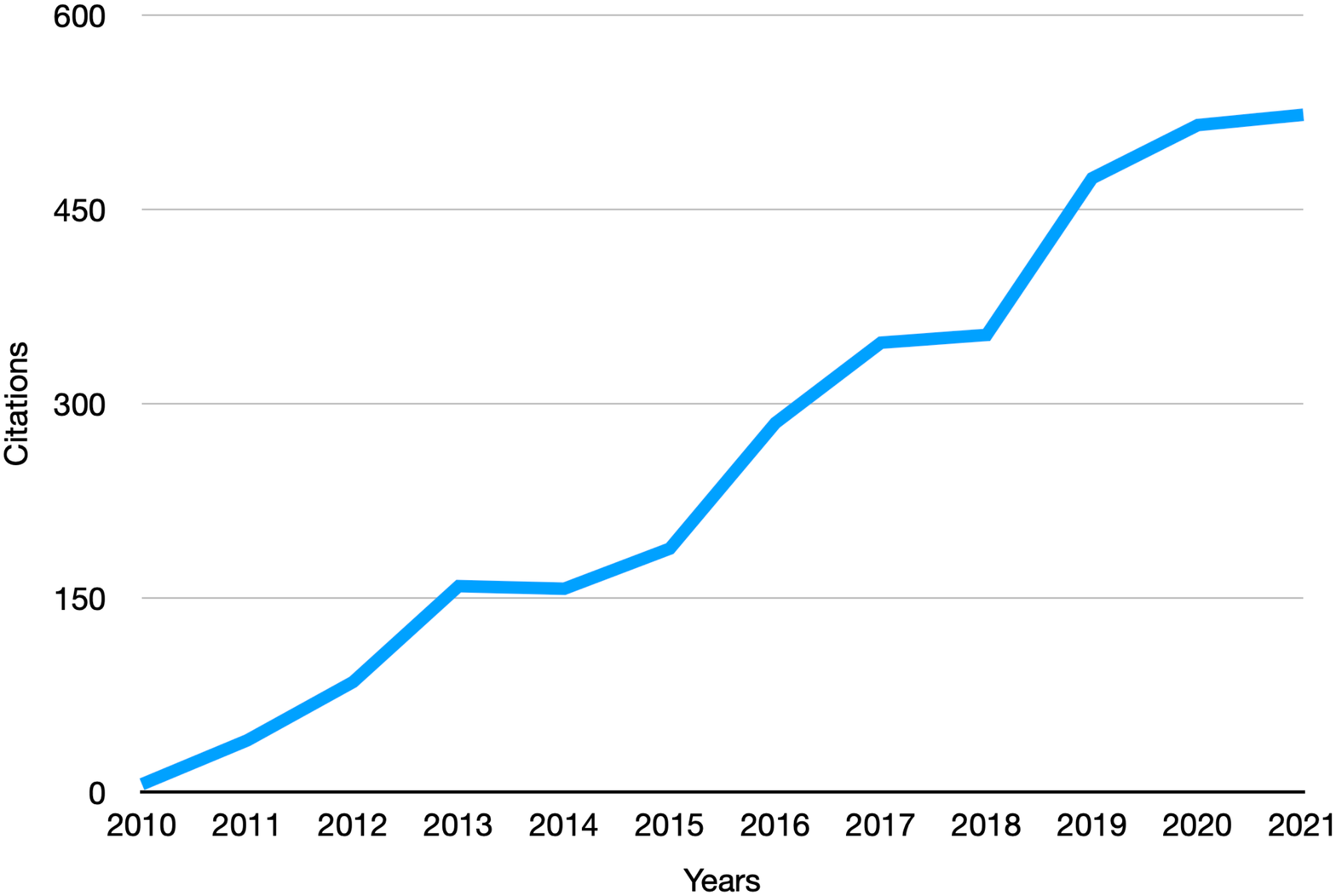}
         \caption{ }
         \label{cites cluster FRBS}
     \end{subfigure}
     \hfill
\caption{ Citations for the published articles during the year 2010-2021: (a) Genetic Fuzzy Systems, (b) Hierarchical Fuzzy Systems, (c) Neuro-Fuzzy Systems, (d) Evolving Fuzzy Systems, (e) FRBSs for Big Data, (f) Interpretability/Explainability in FRBSs, (g) FRBSs for imbalanced data, (h) Cluster centroids as rules in FRBSs.}
\label{cites 2010-2021}
\end{figure}

Table \ref{FS_publishers} shows top conference/journals in FRBSs and the number of articles published in them for GFS during the years 2010-2021. Fig. \ref{GFS publications 2010-2021} shows the trend for the published articles. As it can be the seen, the number of articles are decreasing (with some fluctuations). Fig. \ref{Area-wise publications in GFSs} shows the area-wise contribution in the field of fuzzy systems with the field of computer science having the highest number of articles. Fig. \ref{Citations in the field of GFSs 2010-2021} shows the number of citations for the articles published in GFS.

\textbf{Recent trends in GFS}

In recent years, the researchers in the field of GFSs have focused in the following areas: 
\begin{enumerate}
    \item \textbf{Multi-objective GFS:} There has been many papers in the field of multi-objective GFS where accuracy along with other metrics such as interpretability, complexity are used as objectives. In \cite{su2021multiobjective}, a multi-objective ant colony optimization \cite{dorigo1999ant} has been used to tune IT2FLS parameters for the performance of an Hexapod robot. A multi-objective eFS for intrusion detection has been proposed in \cite{elhag2019multi}. In \cite{gacto2010integration} SPEA2 MOEA was used to obtain pareto fronts between different objectives.
    
    \item \textbf{Hybrid GFS/eFS:} Hybrid form genetic/evolutionary approaches with another methodology are used for learning. In \cite{melin2018hybrid}, a genetic fuzzy neural network approach has been given which combines the benefit of neural networks (NN) with GAs. In \cite{jaafari2019hybrid}, variants of ANFIS with several evolutionary approaches were presented for the prediction of wildfire probability.
    
    \item \textbf{Variants of fuzzy sets in GFS:} There exists several fuzzy sets in the literature such as intuitionistic fuzzy sets, rough set, and type-2 fuzzy sets. Recently, significant focus has been devoted in developing other fuzzy set variants of GFS to capture more uncertainty present in the data. A type-2 evolutionary TSK fuzzy system is presented in \cite{santoso2019t2}. It learns parameters and footprint-of-uncertainty from scratch. A rough-set based variant of GFS was presented in \cite{reddy2020hybrid} for heart disease diagnosis. 
    
    \item \textbf{Applications of GFS:} There has been many areas where GFSs have been making its marks. E.g. the field of power engineering in \cite{mohammadi2017fuzzy}, the GA based fuzzy system reduces upto 46\% of power consumption. Then in \cite{mohamed2014optimized} a GA based optimization of FRBSs water pumping system has been proposed; and in \cite{hameed2021efficient} a system for prediction of heart disease was given.
    
\end{enumerate}
    
\subsection{Hierarchical Fuzzy Systems} \label{FRBS HFS}

Conventional fuzzy rule based systems suffers from the curse of dimensionality. I.e., with the increase in the dimensionality the rule base can become huge. Consider an example of a system which models data with 10 attribute where each attribute can be represented by 5-7 linguistic variables, there are about $5^{10} \sim 7^{10}$ i.e., 10-28 million possible number of rules. In the age of big data, number of attributes are not limited to 10, hence conventional (flat) fuzzy systems are highly infeasible for high dimensional data. Hierarchical Fuzzy Systems were first proposed in \cite{raju1991hierarchical} to overcome the curse of dimensionality. HFS is composed of several low-dimensional fuzzy systems in a hierarchical way. HFSs have also been proved to be Universal approximators \cite{wang1998universal}. Rules in Hierarchical fuzzy systems are grouped into modules (low-dimensional fuzzy systems) as per their roles in the system. Each module computes a partial solution which is further passed onto the next level modules. Although each module is a fuzzy system, it generates significantly less number of rules than a flat fuzzy system. However, this leads to the reduction in the interpretability of HFS as it would not be possible to interpret each intermediate HFS subsystem. 

\begin{figure}
\centering
    \begin{subfigure}[b]{0.26\textwidth}
         \centering
         \includegraphics[width=\textwidth]{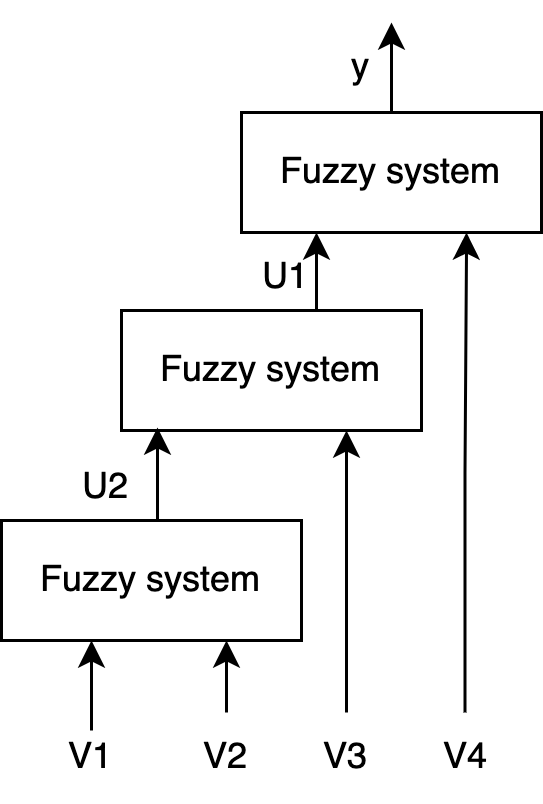}
         \caption{ }
         \label{HFS serial}
     \end{subfigure}
     \hfill
     \begin{subfigure}[b]{0.30\textwidth}
         \centering
         \includegraphics[width=\textwidth]{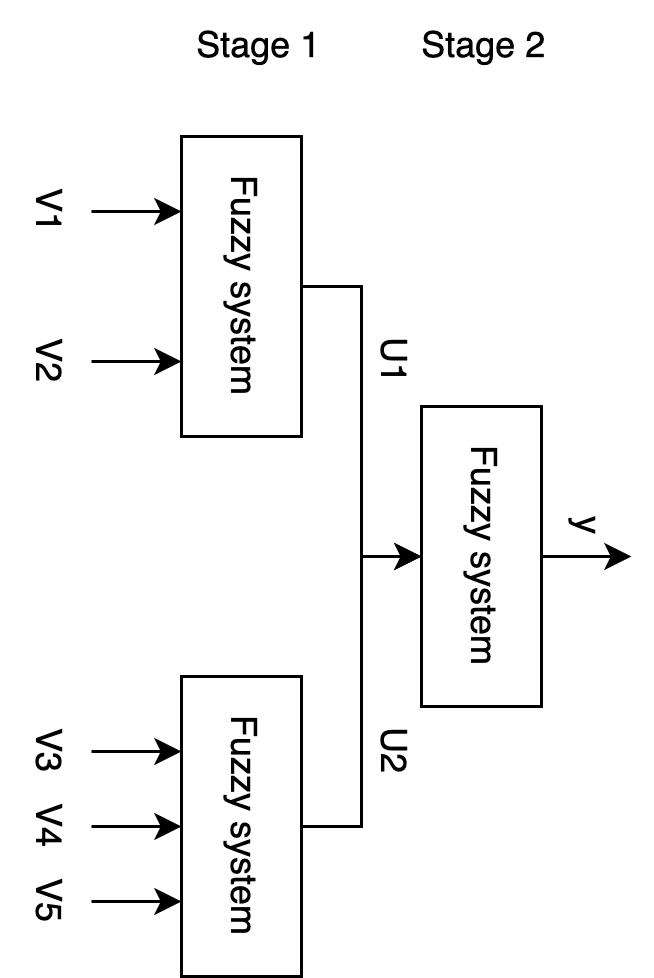}
         \caption{ }
         \label{HFS parrallel}
     \end{subfigure}
     \hfill
     \begin{subfigure}[b]{0.40\textwidth}
         \centering
         \includegraphics[width=\textwidth]{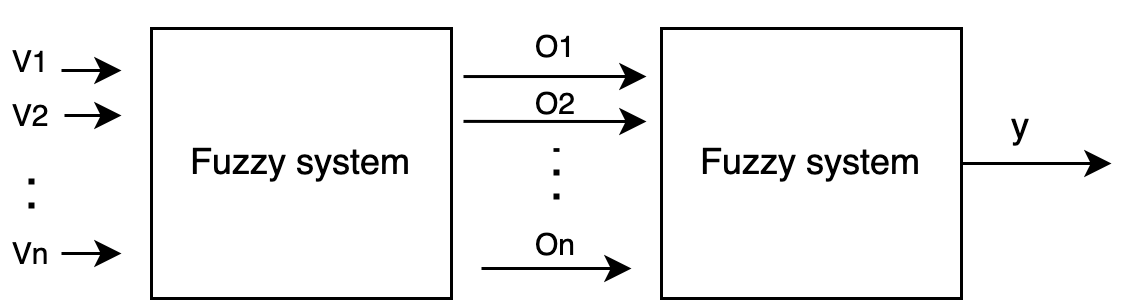}
         \caption{ }
         \label{HFS cascading}
     \end{subfigure}
\caption{Types of Hierarchical Fuzzy System (a) Serial (b) Parallel (c) Cascaded}
\label{HFS_types}
\end{figure}

In general, there are three types of Hierarchical fuzzy systems namely, Serial, Parallel, and Cascaded Hierarchical Fuzzy System. Their structure is given in Figure \ref{HFS_types}. In serial hierarchical fuzzy systems (\ref{HFS serial}), output from the previous modules are fed as one of the inputs in the next module. Here in every stage there is only one fuzzy system. Modules have one input from the previous layer module along with one or more input variable. This process is continued until all the input variables are used.  In parallel (or aggregated) hierarchical fuzzy systems (\ref{HFS parrallel}), the lowest level modules serve as the input for the entire structure. Output of the first level serves as the input of the second level modules. This process continues until the last module. Its output serves as the output of the system. In cascaded hierarchical fuzzy system (\ref{HFS cascading}) each stage is a module which takes all the inputs as parameters. The output of stage 1 module provides the input for stage 2 module. Cascaded HFS are hybrid of FRBSs and Neural networks \cite{duan2001cascaded} but since it uses all of the input variables it loses the benefit of reducing the number of rules. Because of that the field has not progressed much in the past years. 

In the years 2010-2021, several important contributions in the field of Hierarchical Fuzzy Systems have been made. For example, Juang et al. \cite{juang2009hierarchical} provide a particle swarm optimization method for hierarchical fuzzy systems; in Zhang et al. \cite{zhang2014hierarchical} describe a HFS optimized with GAs to develop a robust and accurate traffic prediction system in intelligent transportation problem; Fares et al. \cite{fares2010hierarchical} present a HFS based framework for detecting water mains system; Lopez et al. \cite{lopez2013hierarchical} develop a hierarchical FRBS for classifying imbalanced data. Qu et al. \cite{qu2014cloud} develop a trust evaluation system using HFS was developed for Infrastructure-as-a-service cloud. 

The work done in Wang et al. \cite{wang1999analysis} provides rough analysis and design of HFS. It also highlights that HFSs is a uniform and universal approximator although it can inherit the curse of dimensionality to obtain such general property. In \cite{torra2002review}, a review of HFS from the perspective of complex systems was done. The paper also highlights that if the functions are not decomposable then it may not be possible to generate HFS, but that for some functions HFS is an universal approximator. Review on the approximation capabilities of HFS was done in \cite{zeng2005approximation}. For any continuous function with natural hierarchy, HFS can approximate the system for the desired level of accuracy. A survey paper on HFS which presents the motivation, current trends and open problems in HFS has been presented by Di et al. \cite{di2006survey}. A little work has been done to interpret the intermediate variables in HFS. Magdalena et al. \cite{magdalena2019semantic} shows that HFS only improve the interpretability when it is capable of decoupling the system into subsystems which are interpreted independently.

Table \ref{FS_publishers} shows top conference/journals in FRBSs and the number of articles published in them for HFS during the years 2010-2021. Fig. \ref{HFS publications 2010-2021} shows the trend in number of publications in the years 2010-2021. There has not been much growth in terms of research articles in the field of HFS. Fig. \ref{Area-wise publications in HFSs} shows the areas where HFS papers were published. Most of the papers are in the areas of Computer Science, Engineering and Mathematics. Fig. \ref{Citations in the field of HFSs 2010-2021} shows the citations for the articles published in the years 2010-2021. The increasing trend means that the literature on this topic is increasingly used or constantly improved. 

\textbf{Recent trends in HFS:}

In the recent years, the researchers in the field of HFS have focused in the following areas: 
\begin{enumerate}

    \item \textbf{Interpretable HFS:} Interpretability is one of the main reasons to advocate for the use of FRBSs. Recently many researchers are working towards making interpretable HFS. For example, in \cite{razak2021hierarchical} the authors compare the hierarchical and serial topology of HFS in terms of complexity and interpretation based on Seesaw method, in \cite{razak2020toward} the framework to measure the interpretability with participatory user design for user specific applications was proposed. 
    
    \item \textbf{EA based HFS:} The structure to choose for HFS is a key problem. In cases where relationship between variables are unknown, the search space of possible HFS models is huge. EAs in general are great at finding near optimal solution in huge search spaces. In case of HFS as well, this has been an active area of research. For example, in \cite{zouari2020pso} a PSO-based interval type-2 HFS was proposed for real-time travel route guidance, and in \cite{roy2021hierarchical} a PSO based HFS for reference evapotranspiration prediction was proposed.  
    
    \item \textbf{Type-2 HFS:} Type-2 fuzzy sets are an extension of type-1 to handle more uncertainty present in the data. Researchers have been working towards extending type-1 HFS to the variants of type-2 HFS. E.g., in \cite{wei2021variable} a variable selection method for an interval type-2 hierarchical fuzzy system has been presented; in \cite{jarraya2018multi} the authors introduced a multi-agent architecture for type-2 beta hierarchical fuzzy system which considers different agents to optimize the structure, tune the parameters for improving accuracy and interpretability. 
    
    \item \textbf{Applications of HFS:} There also exists many areas where HFS finds its application. E.g. in \cite{krichen2021autonomous}, HFS has been used to navigate a mobile robot in an unstructured environment; a HFS was proposed in \cite{alrashoud2019hierarchical} for the diagnosis of dengue; in \cite{razak2019measure} HFS has been coupled with software engineering to propose a HFS based complexity measure.
\end{enumerate}

\subsection{Neuro Fuzzy System} 

Neuro Fuzzy Systems (NFS) were first introduced in \cite{jang1991fuzzy}. They present the fusion of FRBSs with Artificial Neural Network (ANN). FRBSs are capable of handling the uncertainty present in the data along with providing interpretable reasoning which can be understood by humans, but FRBSs lacks the ability to learn the rules on its own. On the other hand, ANN is considered a black-box which can learn from the data while not able to provide the reasoning behind the learning. The basic idea behind NFS is to fuse the human-like reasoning ability of FRBSs and learning ability of ANN. There exists two common ways for extracting rules in NFS: cooperative NFS and concurrent NFS. In cooperative NFS, NN computes the FRBSs parameter from training data and FRBSs generates the interpretable rules; in concurrent NFS, NN work together with FRBSs continuously to create the model. NFS can be seen as either a self-adaptive or a dynamic learning system. Automatic tuning of membership function parameters, learning the structure of NFS or both from the data is done in self-adaptive NFS. In dynamic learning NFS, NFS learns parameters or structure from samples continuously rather than learning only initially. However, with the black-box nature of neural network NFS loses its interpretability. 

\begin{figure}
    \centering
    \includegraphics[width=\textwidth]{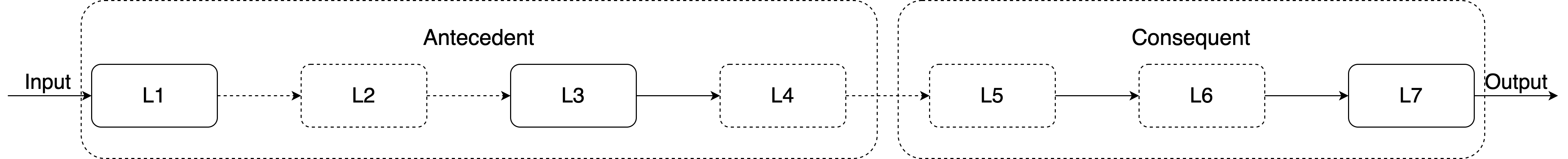}
    \caption{NFS layers, L1-Input layer; L2-membership layer; L3-rule layer; L4-Normalization; L5-Term layer; L6-Extra layer and L7-Output layer.}
    \label{NFS_layers}
\end{figure}

NFS can have at most 7 layers as shown in Fig. \ref{NFS_layers}, Input, Membership, rule, Normalization, Term, Extra, and Output layer. The first four layers (L1-L4) are responsible for tuning and structuring the antecedent of the rule while the others (L5-L7) are used to tune the consequent part. In Fig. \ref{NFS_layers}, and $\dashedrightarrow$ represents partially connected layers, $\rightarrow$ represents fully connected layers. The input is passed through the input layer (L1) without any manipulation. It provides input to either the membership layer (L2) or the rule layer (L3). Input layer is partially connected to membership layer (or rule layer) because each input variable is connected only to their partitioned fuzzy sets. In the membership layer (L2), a problem specific membership function is employed which computes the membership degree corresponding to the fuzzy sets. Usually, a T-norm between the inputs from the input layer or membership layer is done in the rule layer (L3). The partial connection with the previous layers allows the rule layer to learn the structure of the rules, the number of nodes in the rule layer determines the number of rules generated by NFS. The normalization layer (L4) in NFS computes the firing strength of each rule. The term layer (L5) computes the consequent(s) of the rule, and has the same number of nodes as rule and normalization layer. The additional layer (L6) is used to map the input from L5 to a polynomial function which is not very common in NFS research community. The output layer generates the final output of the fuzzy system. In general it is a summation of all the input values from layer L6 (or L5). NFS architecture has at least 3 number of layers (L1-L3-L7 layers) and a maximum number of 7 layers (L1...L7 layers). NFS architecture with higher number of layers offer higher presence of FIS components which also leads to higher efficiency but at the same time may decrease the interpretability of NFS. 

In the literature of NFS, there has been several important contributions during the years 2010-2021 such as the following ones: in Abiyev et al. \cite{abiyev2011type}, a type-2 NFS has been designed for the identification of time-varying systems and authors use clustering methods to equalize time varying methods; in Subramanian et al. \cite{subramanian2013metacognitive}, NFS have been used in the metacognitive framework to develop metacognitive FRBS; in Cervantes et al. \cite{cervantes2016takagi} a TSK based NFS for the identification in non-linear system has been proposed; in Chen et al. \cite{chen2019applying} a hybrid version of NFS with two evolutionary algorithms for the modelling of landslide susceptibility has been proposed; in Feng et al. \cite{feng2018fuzzy} a method to incorporate broad learning in NFS has been proposed for regression and classification problems; the work in Deng et al. \cite{deng2016hierarchical} fuses fuzzy logic with NN in hierachical fashion to create robust classification method.

One of the first review for NFS was done in \cite{nauck1997neuro} where three NFS models were reviewed keeping the model simple and with the same fuzzy semantics. Following the same principle as in \cite{nauck1997neuro}, \cite{nurnberger1999neuro} also reviews the NFS models based on NEFCON-model. In \cite{babuvska2003neuro}, a review of NFS for non-linear system identification has been presented. This review also highlights that modelling of NFS is rather a complex procedure and requires user interaction constantly. The review for various methodologies and applications for NFS during the years 2002-2012 has been done in \cite{kar2014applications}. In \cite{hassan2016optimal}, pros and cons of the optimization techniques using derivative, non-derivative, and hybrid approaches for type-2 NFS has been presented. Recently, a compresive review of the architecture of the NFS along with the review of the optimization approaches has been presented (see \cite{talpur2022comprehensive}). There has been several articles which review the use of NFS for various applications. For example, building energy consumption in Naji et al.\cite{naji2016application}, for estimating power coefficient using NFS in Petkovic et al. \cite{petkovic2013adaptive}, etc.

Table \ref{FS_publishers} shows top conference/journals in FRBSs and the number of articles published in them for NFS during the years 2010-2021. Fig. \ref{NFS publications 2010-2021} shows the trend in number of publications in the years 2010-2021. The research field is seeing increasing number of papers in the field of NFS each year. Fig. \ref{Area-wise publications in NFSs} shows the areas where NFS papers were published, most of the papers are in the areas of Engineering, Computer Science and Mathematics. Fig. \ref{Citations in the field of NFSs 2010-2021} shows the citations for the articles published in the years 2010-2021, the increasing trend means that the literature on this topic is increasingly used or constantly improved. 

\textbf{Recent trends in NFS}

In the recent years, the researchers in the field of NFS have focused in the following areas: 
\begin{enumerate}

    \item \textbf{Explainable NFS:} With the increase in the use of AI/ML in day-to-day life, Explainability/Interpretability is one of the key areas to improve and many researchers are also working towards making explainable/interpretable NFS. In \cite{chimatapu2020hybrid} a deep type-2 FRBSs with the special focus on high dimensional data; in \cite{yeganejou2019interpretable} a NFS which uses CNNs to extract features and then uses fuzzy classifiers to generate interpretable results; and many more. 
    
    \item \textbf{EA based NFS:} Evolutionary Algorithms present an interesting direction as meta-heuristic approach for the optimization of NFS. For example, in \cite{pannu2019multi} PSO has been used for the parameter tuning of NFS to predict the concentration of benzene in the air, in \cite{el2021enhanced} authors uses reptile search algorithm \cite{abualigah2022reptile} for predicting swelling potentiality in Egyptian soil, and in \cite{chen2019applying} PSO was used in NFS for modelling landslide susceptibility.  
    
    \item \textbf{Hybrid NFS:} NFS is a fusion of FRBSs with NN. There has been an increasing interest for other hybrid variants of NFS. For example, in \cite{deng2016hierarchical} the authors proposed a fused hierarchical DNN for classification in image segmentation; in \cite{sumit2019c} fuzzy c-Means clustering has been used to determine the labelling of the attributes and then NFS was employed for the classification in the traffic management. 
    
    \item \textbf{Applications of NFS:} There also exists many areas where NFS finds its application. E.g. prediction of the number of unique online visitors for a journal has been predicted using NFS in \cite{mahmudy2021genetic}, modelling of landslide susceptibility was in \cite{chen2019applying}, and the number of fuzzy rules and membership functions are computed using NFS for regression problem in \cite{juang2011fuzzy}.
\end{enumerate}

\subsection{Evolving Fuzzy Systems} 
 
Nowadays, there is an increasing demand for time-varying systems, systems which can update themselves with the arrival of new data. Data streams are one of the main reasons to look for evolving, adaptive models. Evolving Fuzzy systems (eFS) are a type of FRBSs which can self-adapt the parameters and the structure of the rule with the incoming data stream in online mode. eFS determines whether the incoming data stream can be used to update the parameters, structure, generate a new rule or delete a rule. An on-line clustering between current data and the existing rule is used to check if a new rule will be generated or will be used to update the input space partitioning. A new rule is generated if the new data is some threshold apart from its nearest rule. A rule needs to be deleted before adding a new rule if the rule base already has a predefined maximum number of rules \cite{angelov2002identification}. The rule with the lowest potential is deleted. Potential here represents how much the rule is representative of the data. Consequent of the rule is updated with the incoming data stream. Incoming data can also be an outlier. So in general the consequent is updated in a recursive fashion as done in \cite{de2007line}. Incremental learning is one of the common approaches for training and testing incoming data streams. Incremental learning should take care of the noise and the concept drift with the incoming data. Noise and concept drift represents the change in the relationship between the input vector and the output(s). eFS uses the evolving clustering to detect the concept drift and then update the rule antecedent, and consequent along with managing the number of rules. However, clustering is susceptible to the outliers and noise. Other robust algorithm may present interesting work direction. 

The eFS can be any of the standard FRBS, HFS, NFS which incorporates incremental learning approach to evolve itself with the incoming data stream. In eFS for standard FRBSs, evolving clustering methods are used to update the antecedents; and for addition or removal of the rules in rule base. One such method was proposed in \cite{angelov2010evolving} for TSK-FRBSs. It updates its rule base with the new incoming data using online clustering. The evolvable extension of NFS, are called adaptive or self-evolving NFS methods. In adaptive NFS, models learn to update the antecedent part using evolving clustering methods and simultaneously the weight for the consequent(s). With the incoming data stream, in tree like evolvable structure of FRBSs, a model from a set of neighbour is chosen if the performance of the current model is not optimal as done in \cite{shaker2013evolving}. In \cite{lemos2011fuzzy} an incremental learning approach for linearly evolving tree like FRBSs, which updates its leaves with sub trees based on model selection test to improve the model's performance is presented. 

There has been several important contribution in the field of eFS which have received attention by the research community during the years 2010-2021. For example in \cite{angelov2010evolving}, a multi-input multi-output extension of evolving Takagi-sugeno called eTS+ methodology which is recursive (suitable for data streams) was presented. It allows to shrink or expand the rule base based on the age and potential of the rule and also allows to chose variable selection for incoming data stream; a methodology to deal with concept drift and shift in on-line data stream has been presented in \cite{lughofer2011handling}; in \cite{lin2012identification} a self-evolving recurrent fuzzy neural network for prediction in time-varying systems has been proposed. Unlike TSK model, the methodology considers non-linear relationships between the input variables. In \cite{pratama2013panfis} a model that learns rule bases from scratch, and based on the incoming data stream, can easily add or remove the rules is presented. The contribution of the rule in the output is the major criteria for addition or removal of the rule. Generalized eFS model which considers projection of high dimensional input data into a single fuzzy set has been presented in \cite{lughofer2011handling} this results into more interpretable result. Features in this model can be removed from the rules depending on their relevance, and added in a later stage if needed. A multivariable gaussian eFS model \cite{lemos2010multivariable} which uses multi-variate membership functions for fuzzy sets and uses recursive clustering methods for its robust behaviour has been proposed. There have not been many review articles for eFS, in \cite{baruah2011evolving} a systematic overview for eFS has been presented, they have broadly categorized eFS into two categories: evolvable standard FRBSs and evolvable NFS. Similarly in \cite{ojha2019heuristic}, a brief review of learning and implementation approaches for eFS has been discussed. 

Table \ref{FS_publishers} shows top conference/journals in FRBSs and the number of articles published in them for eFS during the years 2010-2021. Fig. \ref{eFS publications 2010-2021} shows the trend in number of publications in the years 2010-2021. The trend for the number of papers hasn't been consistent but overall there is an average of 45 papers in the field of eFS each year. Fig. \ref{Area-wise publications in eFSs} shows the areas where eFS papers were published, most of the papers are in the areas of Computer Science, Engineering and Mathematics. Fig. \ref{Citations in the field of NFSs 2010-2021} shows the citations for the articles published in the years 2010-2021. The increasing trend for eFS has seen some fluctuations in the recent years, but the interest in the development and use of eFS is growing. 

\textbf{Recent trends in eFS}

In the recent years, the researchers in the field of eFS have focused in the following areas: 
\begin{enumerate}

    \item \textbf{Neuro-eFS:} Evolvable NFS has been one of the prominent approaches in the field of eFS in the recent years where online approaches are used to incorporate data stream for the purpose of tuning or learning the structure of NFS. E.g. in \cite{lin2018self}, a self-evolving interval type-2 NFS has been presented for system identification and control problem. The architecture present learns the optimal structure of NN with the incoming IT2 data. Similarly in \cite{lin2012identification} recurrent approaches along with online structure learning algorithms have been used. 
    
    \item \textbf{Structural Updates in eFS:} The purpose of Evolvable FRBS is to learn the structure, update the rule base with the incoming new data. For example, in \cite{huang2021jointly} a compression layer has been used for the antecendents to eliminate the irrelevant information from the rules which results in higher generalization; in \cite{sa2019structural} a structural evolving approach to improve the accuracy-interpretability trade-off has been presented. This latter study also presents several methodologies to reduce the number of rules in the model. 
    
    \item \textbf{Type-II eFS:} Type-II fuzzy sets are an extension of type-I fuzzy sets whose membership grade itself is a fuzzy set. There has been an increase in interest for type-II variants of eFS in the recent years. E.g., in \cite{lin2018self} a type-2 evolvable NFS has been proposed; in \cite{tung2013et2fis} an evolvable type-II eFS model which considers long-term potentiation if the incoming data stream enhances the previous knowledge and long-term depression if the new data degrades the previous knowledge to update the FRBS model. 
    
    \item \textbf{Applications of eFS:} There also exists many areas where eFS are applied. E.g. for online identification of 1000 different channels for Engine test benches in \cite{lughofer2010data}, for gene data expression in \cite{kasabov2007evolving}, for thermal modelling of power transformers in \cite{alves2021novel}.
\end{enumerate}

\subsection{FRBSs for Big Data} \label{FRBSs big data}

\begin{figure}
    \centering
    \includegraphics[width=\textwidth]{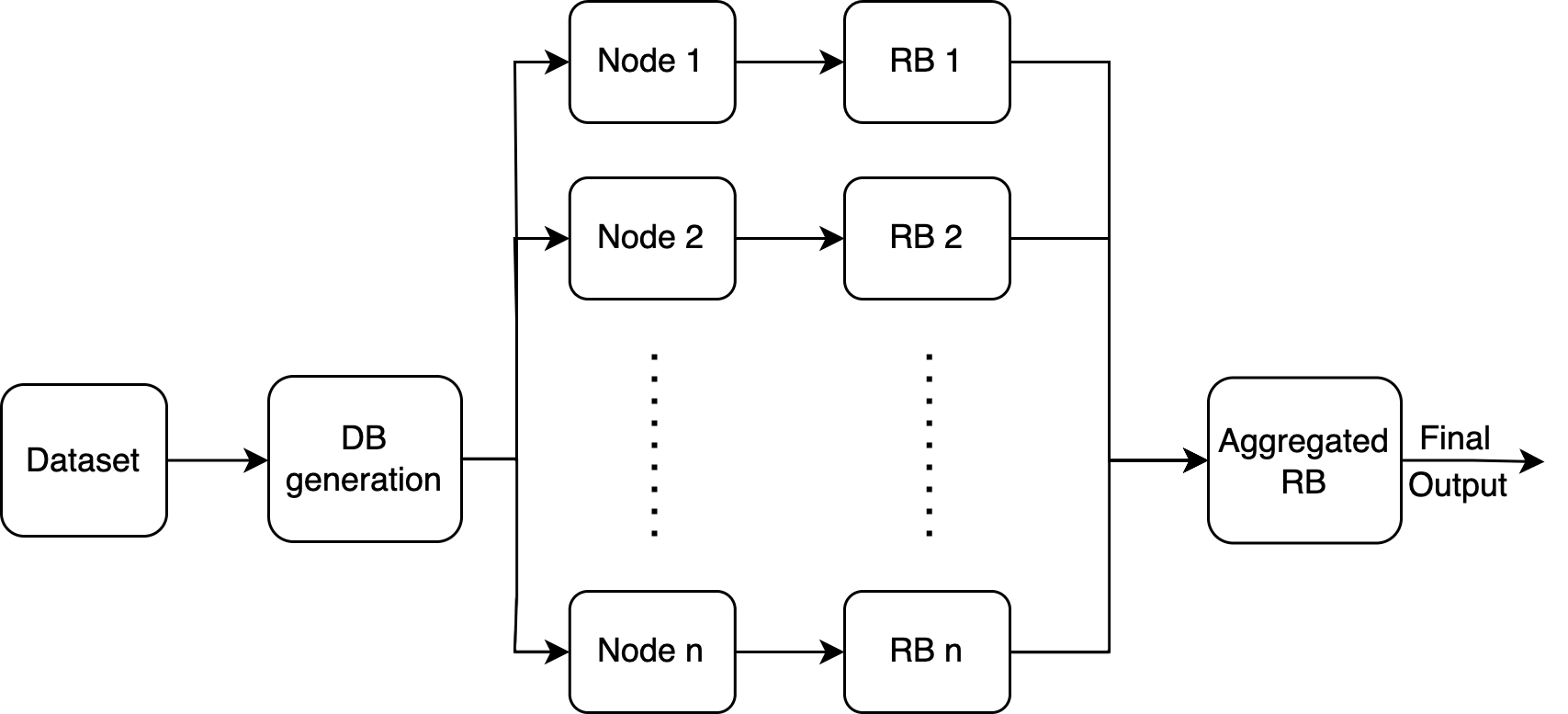}
    \caption{Flow of FRBS with BigData}
    \label{Flow_BigData}
\end{figure}

The amount of data collected in today's world is huge due to internet, mobile devices, social media and many other sources. This huge collection of data is termed as big data. There exists several definitions of big data: some considers big data if has 3Vs to 5Vs \cite{chen2014data} (Volume: huge data, Velocity: high speed of data creation, Variety: data has diversity, Veracity: Data must be reliable and Value: worth of the collected data). Some define big data as the data which can not be processed in a single machine \cite{elkano2018chi}. FRBSs are known for generating interpretable models but FRBSs struggle with the scalability issue, so it does not perform well for big data. To solve this issue, researchers have been working over generating distributed, parallel models in the past decade, hence the number of publication in the field has significantly increased (see fig. \ref{Big data publications 2010-2021}). The basic idea is to use distributed computing, where data is distributed among several computing nodes. At each node, the splitted data is used to build FRBS model and the output generated from each node is aggregated to get the final output as shown in Fig \ref{Flow_BigData}. The problem of interpretability with high dimensional data in case of big data is still a major drawback of fuzzy systems in big data.

Mapreduce by Dean et al. \cite{dean2008mapreduce} is one of the common approaches for distributing the large data into a cluster of nodes. Mapreduce has two primary steps, Map and Reduce. First, the Map function distribute the data into nodes and computes some intermediate result then Reduce aggregates the intermediate result into the final output of the model. Chi's FRBS \cite{chi1996fuzzy} is compatible with the Mapreduce paradigm, the first model which uses Chi's method with Mapreduce called CHi-FRBCS-BigData was given in \cite{del2015mapreduce}. Initially, fuzzy partition of each feature in the database is created based on the level of chosen granularity. Then the whole dataset is distributed among the n computing nodes. In the second stage, each of the nodes independently computes its rule base based on Chi's or any rule learning method. In the third stage, the rule base from each node is aggregated and in case of conflict the rule with the higher rule weight is considered. The initial computed DB and final rule base results in the KB for this type of models. 

There has been several important contributions to the field of fuzzy systems with Big Data which have received attention in the research community during the years 2010-2021. Del et. al \cite{del2015mapreduce} describe above is one example. This is about an FRBS model for the classification in big data context using the Mapreduce approach on the Hadoop platform. This work was further extended using cost-sensitive learning for imbalanced dataset in \cite{lopez2015cost}. Both approaches does not lead to the same rules if Chi's original algorithms was used. This drawback was removed in \cite{elkano2018chi}. In \cite{segatori2017distributed} a distributed fuzzy classifier which employs frequent pattern for pruning unnecessary rules is introduced . A brief overview of Chi-FRBS approaches for big data along with granularity analysis has been given in \cite{fernandez2017fuzzy}. Their results suggest that increasing granularity increases utility, but having too high granularity ($>$5 in their experiments) can also cost loss of utility. The work in \cite{fernandez2016view} highlights that fuzzy systems for big data are still in its early stage and provides a brief review on the following key areas: classification, association rule mining and clustering for big data. In \cite{wang2017overview}, an overview of fuzzy techniques has been presented where fuzzy approaches have been used to handle uncertainty and for modelling purposes. They also highlight that most of the current fuzzy work in Big Data only consider volume as their definition of big data. 

Table \ref{FS_publishers} shows top conference/journals in FRBSs and the number of articles published in them for Big data in FRBSs during the years 2010-2021. Fig. \ref{Big data publications 2010-2021} shows the trend in number of publications in the years 2010-2021. The number of publications in the field haven't been consistent. Fig. \ref{Area-wise publications Big Data} shows the areas where papers in FRBS in big data were published. Most of the papers are in the areas of Computer Science, Engineering and Mathematics. Fig. \ref{Citations in the field of Big Data 2010-2021} shows the number of citations for the articles published in the years 2010-2021. The field was introduced recently and the trend suggests the research in FRBSs for Big Data has been receiving constant attention from the research community. 

\textbf{Recent trends in FRBSs for Big Data}

In the recent years, the researchers in the field of FRBSs for Big Data have focused in the following areas. 
\begin{enumerate}

    \item \textbf{Neuro FRBS-BigData:} In the present stage, neural networks is one of the key machine learning areas where progress is being made constantly. FRBS-BigData is no different. There has been several recent articles which present NFS for big data such as the following. In \cite{de2017usnfis}, multi-layer NFS with time varying learning rate for Big data is presented, in \cite{zhang2020hierarchical} a two-stage optimization for privacy-preserving Hierarchical-NFS has been presented. In first stage, parameters are trained using distributed K-means and AO algorithm is used for coordination among nodes at higher levels.
    
    \item \textbf{Structural Updates in FRBS-BigData:} Many researchers in FRBS-BigData are also looking to optimize the rule complexity and the running time of the model such as in \cite{iniguez2018improving} the author present a support based filtering of rules instead of rule weight based filtering for the classification task in FRBS-BigData. This system results in an improved accuracy and running time. In \cite{aghaeipoor2021ifc} an interpretable FRBS which tries to reduce the number of rules as well as the number of features in a rule has been presented.
    
    \item \textbf{Applications of FRBS-BigData:} There also exists many areas where FRBS-BigData finds its application. E.g. in \cite{hu2021fuzzy} brain disease prediction using medical image processing has been proposed, learning model for supply chain management using ANFIS for big data has been proposed in \cite{bamakan2021di}.
\end{enumerate}

\subsection{FRBSs for Interpretability/Explainability}

FRBSs are widely used because of their ability to provide human understandable knowledge in the form of linguistic rules against their black-box counterparts. FRBS models are interpretable by design which provides the answer to what will be the output of the model and why. The interpretable structure of the FRBSs can be represented in several forms such as: TSK, Mamdani and many others as defined in section \ref{FRBS}. A model which is highly interpretable but has low utility will never be used, so interpretability alone is not sufficient. Two generic types of fuzzy modelling approaches are famous, Linguistic fuzzy modelling (LFM) where the primary objective is to have high semantic interpretabilty while Precise fuzzy modelling (PFM) focus on highly accurate fuzzy models. There are many aspects in the design of FRBS which improve the interpretability and accuracy of the model. E.g. the granularity level, the membership function and many more. Interpretability of FRBSs is not only associated to classical FRBS but also to HFS, NFS, GFS. HFS divides the large FRBS system into subsystems to reduce the number of rules. Each of the subsystem generates an intermediate variable which is used as input in the next level. Magdalena et. al \cite{magdalena2018hierarchical} highlights that the use of intermediate variables leads to the black-box FRBS. A comprehensive study on the interpretability of the HFS is yet to be done. In GFS, the evolutionary algorithms are used for the learning purpose in the KB \cite{cordon2011historical}; and in NFS, the NNs are used for learning purposes \cite{paiva2004interpretability}. 

There is a general belief that there exists a trade-off between accuracy and interpretability, even though it may not be necessarily true. There exists two approaches in the accuracy-interpretability trade-off. Some try to improve the accuracy of the model while maintaining a good interpretability of the system while other methods try to make the model more interpretable while maintaining high accuracy. The accuracy of an interpretable-FRBS is further improved by updating the rule structure which encompasses the granularity, choosing the MFs and their parameters, limiting the number of rules and the number of variables in each rule etc; and improving the learning process using additional components (such as GFS). Interpretability can further be improved by reducing the number of features in the rules; reducing the number of rules (one such approach is to merge close enough rules); simplifying the output of the FRBSs. These approaches focus on the sequential improvement of accuracy for LFM or improving interpretability for PFM. Some researchers also considers these in an interleaved way. E.g., in \cite{knapp2001refine}, first an LFM is constructed then its accuracy is improved by generating large fuzzy partitions which generally result in a huge rule base. Then rules are merged again to improve the interpretability of this model. This can be extended till the desired result is obtained. Since interpretability and accuracy are generally conflicting objectives, some researchers focus on multi-objective EA such as SPEA, NSGA-II and many other EAs. In case of big data, the rules generated in FRBS may not be highly interpretable as the number of variables in the rule may be quiet large to obtain the accurate model. Recently, the interpretability problems in FRBS for big data have also been highlighted \cite{hullermeier2015does}. In \cite{varshney2022designing}, 1-dimensional interpretable rules have been presented. A critical analysis of interpretability for Mamdani or TSK type FRBSs has been presented in \cite{mendel2021critical}.  

There have been several important contributions in the interpretability field of FRBSs. A field that has received attention by the research community during the years 2010-2021. E.g., in \cite{pulkkinen2009dynamically} a constrained multi-objective GFS which uses decision trees to initialize the rulebase that results in less rules with less variables due to a reduced search space. A generic framework to evaluate and design interpretable FRBS has been given in \cite{guillaume2011learning}. In \cite{cpalka2014new}, a NFS that focuses on accuracy along with high interpretability (as defined in \cite{gacto2011interpretability}) for non-linear systems has been given. Hagras et. al \cite{hagras2018toward} presents an overview and stresses the need of understandable and explainable AI. There has not been much work regarding interpretability of HFS and there is a lack of interpretable measures for HFS. In \cite{razak2017interpretability} an index for HFS has been presented. A detailed review of interpretability in FRBSs until 2010 has been done in \cite{gacto2011interpretability}. They also highlight that there is no comprehensive measure that quantifies interpretability. Interpretability has to be seen from different characteristics such as number of rules, size of rule, granularity etc. The work in \cite{lughofer2013line} highlights the open issues, new development, trends and achievement for eFS/GFS. A review for Mamdani-type FRBSs in the context of GFS has been presented in \cite{cordon2011historical}. Another review which focuses on the information granule from the the perspective of interpretability-accuracy trade-off has been given in \cite{ahmed2017knowledge}.

Table \ref{FS_publishers} shows top conference/journals in FRBSs and the number of articles published in them for interpretability/explainability in FRBSs during the years 2010-2021. Fig. \ref{XAI publications 2010-2021} shows the trend in number of publications in the years 2010-2021. The trend for the number of publications in the field has not been consistent, although from 2014 the number of publications are increasing. Fig. \ref{Area-wise publications XAI} shows the areas where papers in FRBS in big data were published. Most of the papers are in the areas of Computer Science, Mathematics and Engineering. Fig. \ref{Citations in the field of XAI 2010-2021} shows the citations for the articles published in the years 2010-2021. The linearly increasing trend suggests the work done is being used or improved constantly.

\textbf{Recent trends in interpretability/explainability in FRBSs:}

In the recent years, the researchers in the field of interpretable FRBS have focused in the following areas: 
\begin{enumerate}

    \item \textbf{Interpretability of FRBS in Big data:} The whole idea of fuzzy systems revolves around the idea that they are not black-box models. In contrast, they follow the interpretable by design paradigm. But as highlighted in \cite{mendel2021critical}, it is very difficult for humans to correlate more than two variables in a rule, let alone 10,20 or more variables. In \cite{aghaeipoor2021ifc}, an improvement for horizontal and vertical size of the rule using different filtering approaches, a density based filtering for baseline rule base, considering only the important linguistic variable for interpretable rule, heuristic rule selection for reducing the total number of rules. 
    
    \item \textbf{Interpretability of Hierarchical Fuzzy Systems:} The interpretability of HFS has been linked to reducing the complexity of FRBS but at the same time it uses intermediate variables, which raises the question: reducing the number of rules at the cost of intermediate variables (interpretable or not) leads to higher or lower interpretability?. In the recent years many researchers have started looking for this question. In \cite{magdalena2018hierarchical} not all HFS are interpretable, and the interpretability of the HFS depends significantly on the the choice of intermediate variables. In \cite{razak2017interpretability} the lack of indices to measure interpretability in HFS has been highlighted.
\end{enumerate}

\begin{sidewaystable}
\centering
    \begin{tabular}{cccc}
        \hline
         Fuzzy System & Advantage & Disadvantage & Applications  \\
         \hline
         \hline
         GFS & \begin{tabular} {@{}c@{}c@{}c@{}}Works Independent of data\\ Can handle Vague, inconsistent data \\ Near optimal performance \\ Can be used for tuning and searching \end{tabular} & \begin{tabular} {@{}c@{}c@{}c@{}}Slow learning rate\\ Lacks comparative analysis \\ Many objective functions \\ rarely considered \end{tabular} & \begin{tabular} {@{}c@{}c@{}c@{}} Fuzzy logic controllers \\ Classification, Data Mining \\ Medical diagnosis \\ Forecasting \end{tabular} \\
         \hline
         HFS & \begin{tabular} {@{}c@{}c@{}} Less number of rules \\ Handles high-dimensional data \\ Computationally efficient \end{tabular} & \begin{tabular} {@{}c@{}c@{}} Loose interpretability  \\ tuning rulebase \& architecture \\ can be computationally costly \end{tabular} & \begin{tabular} {@{}c@{}c@{}} Control Systems \\ Feature selection \\ Power Systems \end{tabular} \\
         \hline
         NFS & \begin{tabular} {@{}c@{}c@{}c@{}} No prior knowledge required \\ fast learning \\ Multiple layers to approximate \\ any learning function \end{tabular} & \begin{tabular} {@{}c@{}c@{}} Require structure \\ Loose interpretability in \\ high-dimensional data \end{tabular} & \begin{tabular} {@{}c@{}c@{}} TIme-series data \\ Fuzzy rule generation \\ Healthcare \end{tabular} \\
         \hline
         eFS & \begin{tabular} {@{}c@{}c@{}} Handles real-world scenarios \\ Memory efficient \\ Incrementally learns global model \end{tabular} & \begin{tabular} {@{}c@{}c@{}c @{}} less flexibility with new rules \\ over-fitting in rule-merging \\ Large changes in input/output \\ domains can not be easily resolved  \end{tabular} & \begin{tabular} {@{}c@{}c@{}} Gene expression modelling \\ Real-time decision making \\ On-line identifications \end{tabular} \\
         \hline
         Big data & \begin{tabular} {@{}c@{}c@{}} Distributed paradigm \\ can handle huge volume of data \\ Cost effective \end{tabular} & \begin{tabular} {@{}c@{}c@{}} Challenge with high velocity \\ and variety in data \\ interpretability is a challenge\end{tabular} & \begin{tabular} {@{}c@{}c@{}} Medical imaging \\ Supply chain \end{tabular} \\
         \hline
         Interpretability & \begin{tabular} {@{}c@{}} Generates human-understandable rules \\ Important for legislations like GDPR \end{tabular} & \begin{tabular} {@{}c@{}c@{}} High interpretability constraints \\ Utility-Interpretability trade-off \\ No clear definition or assessment \end{tabular} & \begin{tabular} {@{}c@{}c@{}} XAI \\ Robotics \\ Control Systems \end{tabular} \\
         \hline
         \hline
    \end{tabular}
    \caption{Advantages, Disadvantages and Applications of various types of Fuzzy systems}
    \label{Advantages_disadvantages}
\end{sidewaystable}

\subsection{Other areas:}

Apart from the above mentioned areas, FRBSs have also contributed in the areas discussed in this section.

\subsubsection{FRBSs for imbalanced datasets} 

Real-world problems generally have imbalanced data. Imbalanced datasets are the datasets which have unequal distribution of the output classes. E.g., in spam detection, the number of spam emails are significantly less than the number of legitimate email. Numerous approaches have been proposed in the literature to handle imbalanced datasets. They can be broadly categorized into three categories: Data-level approaches where the training data is modified to obtain the relatively balanced dataset; algorithm level approaches where existing methodologies are modified to improve the classification for minority class; and cost-sensitive approaches where there is a higher penalty associated with the misclassification of the minority class. In the recent years, FRBSs are seen as an effective approach for classification in imbalanced settings where many researchers consider the use of synthetic minority oversampling technique (SMOTE) in data-level approaches. E.g. fernandez et. al \cite{fernandez2009influence} choose two neighbouring points at random and generates a synthetic point from the combination of the two; for algorithm level approaches in FRBSs, algorithmic changes such as in \cite{fernandez2009hierarchical} which presents a hierarchical FRBCSs to generate fine granularity between majority and minority classes using genetic rule selection; and for cost sensitive approaches the cost are included during evaluation of the model to favor the minority classes such as in \cite{ishibuchi2005rule}. Class imbalance in big data has also become a prominent area. E.g., in \cite{lopez2015cost} a cost sensitive FRBS for imbalanced big data has been considered, in \cite{fernandez2017chi} a rule selection method in a spark environment for imbalanced datasets in big data FRBSs has been presented. In \cite{fernandez2016view}, a review of eFS approaches which focuses on classification using imbalanced datasets has been presented. Table \ref{FS_publishers} shows top conference/journals in FRBSs and the number of articles published in them for imbalanced dataset in FRBSs during the years 2010-2021. Fig. \ref{imbalance publications 2010-2021} shows the number of publication in the FRBSs for imbalanced data. The trend has been inconsistent but the research in this direction is still active and has been receiving attention from the research community as can be seen from Fig. \ref{cites imbalanced FRBS}. In the the past years, much of the research has been focused over eFS or Big data for FRBS. 

\subsubsection{Cluster centroids as rules in FRBSs}

In FRBSs, the number of rules can be quiet large. Determining the maximum number of rules can be very problem dependent. In the recent years, clustering has gained attention for applications (or areas) which requires a fixed number of rules. For a user defined number of rules c, the data is clustered into c clusters, where each cluster centroid represents a rule of the fuzzy system. There has been many clustering approaches applied to a variety of fuzzy systems. E.g., fuzzy c-Means has been used in \cite{kerr2020generating} to generate hierarchical structured fuzzy rules, in \cite{nguyen2015hybrid} rule structure was initialized using the possibilistic clustering algorithm for NFS; subtractive clustering has been used in \cite{pham2012learning} to initialize rules for TSK fuzzy systems. These systems generally use a clustering algorithm (a type of c-Means clustering) to model any type of the fuzzy system. Table \ref{FS_publishers} shows top conference/journals in FRBSs and the number of articles published in them for initializing rules using cluster centroids in FRBSs during the years 2010-2021. Fig. \ref{cluster publications 2010-2021} shows the number of publication in FRBSs that use cluster centroids as initial fuzzy rules. The trend has been inconsistent. In general there has been roughly 30 papers each years on the use of cluster centroids in the various types of fuzzy systems. Fig. \ref{cites cluster FRBS} shows the number of citations received for the papers published in the field during the years 2010-2021. The increasing trend suggests that the work is continuosly receiving attention from the research community. 

\section{Open problems in FRBSs:} \label{open problems}

In the past 40-50 years, FRBSs have been successfully used in various areas. Nevertheless there are still many open problems which can present opportunities for the FRBSs research community:

\begin{enumerate}

    \item Ethical challenges: With the increase of AI in day-to-day life, ethical challenges such as privacy, fairness, interpretability with the AI models must be handled immediately. With the increase in the number of publicly available datasets, the need to avoid disclosure of sensitive information is also increasing. Very few worls in the fuzzy community are concerned with the ethical concerns of AI.  In general, there exists a trade-off between utility-interpretability, utility-privacy, utility-fairness but further study is needed to reflect upon the relationship between data privacy, fairness and interpretability. FRBSs can be a candidate model to study the relationship between interpretability, fairness and data privacy.
    
    \item Representation of uncertainty: Fuzzy sets are used to represent the uncertainty of the data. They are the interpretable variables in FRBSs. In the current literature, the focus has been given to classical fuzzy sets and type-II fuzzy sets but there exists many other fuzzy sets such as Hesitant fuzzy sets \cite{torra2010hesitant}, rough sets \cite{dubois1990rough} and so many others. These sets provide the opportunity either enhance the uncertainty representation or consider multiple experts opinion. 
    
    Fuzzy sets use membership functions for uncertainty representation. There exists different families of membership function but there is not much work which compares the impact of different types of membership functions on the performance of FRBSs. E.g. in \cite{mendel2021critical}, authors have suggested that triangular and trapezoidal MFs are more suitable to XAI than Gaussian MFs. MFs also influence the interpretability of FRBS reasoning, FRBSs automated efficient choice, and partitioning for MFs can be an important factor for FRBSs. 
    
    \item Big Data: With the amount of data being produced in today's world, researchers should consider model's performance in the big data context. In the recent years, researchers in classical FRBSs model have started working in the field of classification for big data (e.g. \cite{elkano2018chi}) but there has not been much work in other machine learning aspects. Not much focus has been given for big data in other types of FRBSs such as GFS, NFS and HFS. Genetic/evolutionary algorithms can help in parallelization and improving the learning rate for the MapReduce framework. Further study in this area can provide fast and good solutions. The complexity of NFS increases with big data which will require time consuming and costly solutions. An interesting direction can be distributed/federated approaches for hybridization of NNs and FRBSs. This approach may present robust and fast solutions. Even though the motivation for HFS is its ability to overcome the curse of dimensionality, there hasn't been any significant work which discusses the application of HFS in the big data context.
    
    Another big data problem which is prominently present in the FRBSs community is the consideration of only volume for big data. I.e. big data comprises the 3-5 Vs, as explained in section \ref{FRBSs big data}. Nevertheless, in general researchers have used huge data volume as big data. Consideration of other Vs in FRBSs can provide important insights in the respective fields. 
    
    \item Evolving nature of Algorithms: In real-world applications, data is often in streaming nature. I.e. data distribution keeps changing with time. This requires continuous development of the developed model. Generally, evolving FCM is used to determine the age and the rule based on the incoming data stream. Evolvable FCM is sensitive to the outliers and noises. A more robust approach in the evolving fuzzy systems may significantly enhance its performance. Apart from this, more focus should be given for the selection of features, input, and the aggregation of rule in eFS. Although, eFS considers streaming data, none (or few) of the papers has considered the impact of high (or low) velocity of incoming data on the performance of eFS. eFS also does not provide the flexibility architecture of the rule and system e.g. Hierarchical or distributed architecture.
    
    \item Aggregators in FRBSs: Aggregation in FRBSs can be used for merging the fuzzy rules and reduces the number of rules in the system, and to get the firing strength of a rule. In general, neighbour rules which after merging does not affect accuracy (much) are merged using an aggregation operator for each feature. The features in a rule of FRBSs are connected through a conjunction which is usually modeled through a t-norm. There exists several other aggregation operator such as OWA \cite{yager1988ordered}, Choquet \cite{choquet1954theory} and Sugeno \cite{sugeno1974theory} integrals and many more. It is relevant to consider the aggregation functions that can have a significant impact on the performance of the FRBSs. Different aggregation operators on various types of FRBSs.
    
    \item Interpretability/Explainability: Interpretability is a key aspect of FRBSs. Nevertheless in the last few years, critical analysis of fuzzy systems with respect to interpretability has gained attention from researchers. As suggested in \cite{mendel2021critical}, if a rule contains more than 2 antecedents it is very hard for the human mind to understand. Further study in this area regarding the loss of accuracy with 2-3 antecedents in the rule may be needed. As mentioned in Section \ref{FRBS HFS}, the interpretability of HFS needs to be further analysed. Apart from this, focus must be given for new interpretability assessment indices to deal with the complex FRBSs structure. 

    \item Model optimization: Optimization of Mamdani and TSK FRBSs is still an open problem. The choice for membership functions, number of rules, defuzzifiers are still done manually, and the automation of this process is an interesting future work direction. Meta-heuristic approaches can be used for this goal. 

    \item Application in Robotics: Robotics becomes more and more complex and challenging with time. Fuzzy sets and systems can be used to deal with the uncertain information available to the robots. This can be very helpful for robots in their need to navigate in dynamic and unpredictable environment. 
    
    \item Deep Fuzzy multi-layer network: In the last few years, significant focus has been given to deep fuzzy neural networks. They incorporate a fuzzy layer in the DNN, and, thus, it offers interpretability to the otherwise black-box model. Due to the deep architecture along with high number of parameters to tune, DFNN has very high computational cost for big data, distributed or parallel. Most of research articles consider gradient descent to optimize the parameters but there exists many other approaches which can result in a better performing DFNN, further analysis in this area is needed.
\end{enumerate}

\section{Limitations}

This article presents a brief overview of the developments in the different types of fuzzy systems during the years 2010-2021. As discussed above, Fuzzy Systems have been used in various areas and a huge number of papers are published every year, discussion or inclusion of this many papers is quiet difficult. So, this review article suffers from the following limitations:

\begin{enumerate}
    \item This article provides a brief overview, key papers which received attention in research community and the current trends in the various areas of FRBSs but at the same time it does not cover all the papers, their approaches, and their classification as done in \cite{shihabudheen2018recent}. 
    
    \item This article does not dive deep into the architecture and methodology of the each type of FRBSs.
    
    \item FRBSs has been in the literature for quiet some time and literature for each type of FRBSs is huge, so coverage of each article published in each various types of FRBSs would be beyond the scope of this article.
    
    \item Since use cases of each type of FRBSs are different, this article does not provide the comparison between each type of FRBSs.
\end{enumerate}

\section{Conclusion}

This article provide answers to the research questions asked in Section-I. Fundamental concepts of FRBSs have been highlighted by presenting an overview for the various types of classical fuzzy rule based systems and variants of FRBSs. The six variants of classical FRBSs namely Mamdani FRBSs with input-output scaling, DNF FRBSs, TSK FRBSS, Approximate FRBSs, MIMO FRBSs and FRBCSs. The paper identifies eight variants of FRBSs which receive most of the attention from FRBSs community namely, genetic/evolutionary fuzzy system, hierarchical fuzzy systems, neuro fuzzy systems, evolving fuzzy systems, advances in FRBSs for big data, FRBSs for interpretability/explainability, FRBSs for imbalanced datasets and FRBSs that uses cluster centroids as rules. For each of the topics, a brief overview, review papers, trending papers and trending areas related to each topic during the year 2010-2021 are presented. The articles present in the scopus database have been used to get the publishing statistics related to each of the topic for RQ2. Based on the analysis of publishing statistics (in Section \ref{lit-review}) related to each variant of FRBS, the paper suggests that the most active trend nowadays in the field of FRBSs lies in NFs, big data and interpretability of FRBSs. Significant reduction in the research community for GFS can be observed. It is also clear that hybridization of two or more types of models are gaining more and more popularity. The open research areas in FRBSs are highlighted in Section \ref{open problems} which presents the potential future direction for FRBSs out of which Ethical challenges in FRBSs, other Vs of Big data, Model optimization and applications in robotics presents an interesting future direction.

\printbibliography

\vspace{1.0cm}
\textbf{Statements \& Declarations}

\textbf{Acknowledgement: } Authors are thankful to the editors and the reviewers for their valuable comments which have helped a lot in improving the work. This work was partially supported by the Wallenberg Al, Autonomous Systems and Software Program (WASP) funded by the Knut and Alice Wallenberg Foundation.

\textbf{Author Contribution: } Ayush K. Varshney: Conceptualization, Methodology, Formal analysis, Validation, Visualization, Writing - original draft. Vicen\c{c} Torra: Conceptualization, Formal analysis, Validation, Writing – review \& editing, Resources, Supervision. 

\textbf{Data availability: }Data sharing not applicable to this article as no datasets were generated or analysed during the current study.

\textbf{Declaration of competing interest: }The authors declare that they have no known competing financial interests or personal relationships that could have appeared to influence the work reported in this paper.
\end{document}